\documentclass[10pt,journal,compsoc]{IEEEtran}
\usepackage{textcomp}
\usepackage{color}
\usepackage{graphicx}
\usepackage{amsmath}
\usepackage{amssymb}
\usepackage{algorithm}
\usepackage{algorithmic}
\usepackage{verbatim}
\usepackage{multicol}
\usepackage{amsmath}
\usepackage{enumerate}
\usepackage{tabularx}
\usepackage{subfigure}
\usepackage{graphicx}
\usepackage{booktabs}
\usepackage[labelfont=bf,format=plain,justification=raggedright,singlelinecheck=false]{caption}
\usepackage{hyperref} 
\hypersetup{hypertex=true,
            colorlinks=true,
            linkcolor=red,
            anchorcolor=red,
            citecolor=red}
\newtheorem{theorem}{Theorem}
\newtheorem{lemma}{Lemma}
\newtheorem{proof}{Proof}[section]
\newtheorem{assumption}{Assumption}
\newtheorem{definition}{Definition}
\newcommand{\etal}{\emph{et al.} }
\newcommand{\w}{\mathbf{w}}
\newcommand{\s}{S_a^{(t)}}
\newcommand{\D}{\mathcal{D}}
\newcommand{\B}{\mathcal{B}}
\newcommand{\p}{\mathcal{P}}
\ifCLASSOPTIONcompsoc
  \usepackage[nocompress]{cite}
\else
  \usepackage{cite}
\fi
\ifCLASSINFOpdf
\else
\fi
\begin{document}
%
\title{A Decentralized Federated Learning Framework via Committee Mechanism with Convergence Guarantee}
%
%
%
%

\author{Chunjiang Che,
 Xiaoli Li,~\IEEEmembership{Student Member,~IEEE,} \\
 Chuan Chen,~\IEEEmembership{Member,~IEEE,}
 Xiaoyu He,
 and~Zibin Zheng, ~\IEEEmembership{Senior Member,~IEEE}
 \thanks{Chunjiang Che, Xiaoli Li, Chuan Chen, and Xiaoyu He are with the School of Computer Science and Engineering, Sun Yat-sen University, Guangzhou, China. Zibin Zheng is the School of Software Engineering, Sun Yat-sen University, Zhuhai, China. e-mail: \{chechj, lixli27\}@mail2.sysu.edu.cn, \{chenchuan,
hexy73, zhzibin\}@mail.sysu.edu.cn.
 Corresponding author: Chuan Chen.}
}

%
%

\markboth{Journal of \LaTeX\ Class Files,~Vol.~14, No.~8, August~2015}%
{Shell \MakeLowercase{\textit{et al.}}: Bare Demo of IEEEtran.cls for Computer Society Journals}
%



\IEEEtitleabstractindextext{%
\begin{abstract}
Federated learning allows multiple participants to collaboratively train an efficient model without exposing data privacy. However, this distributed machine learning training method is prone to attacks from Byzantine clients, which interfere with the training of the global model by modifying the model or uploading the false gradient. In this paper, we propose a novel serverless federated learning framework \textit{Committee Mechanism based Federated Learning} (CMFL), which can ensure the robustness of the algorithm with convergence guarantee. In CMFL, a committee system is set up to screen the uploaded local gradients. The committee system selects the local gradients rated by the elected members for the aggregation procedure through the selection strategy, and replaces the committee member through the election strategy. Based on the different considerations of model performance and defense, two opposite selection strategies are designed for the sake of both accuracy and robustness. Extensive experiments illustrate that CMFL achieves faster convergence and better accuracy than the typical Federated Learning, in the meanwhile obtaining better robustness than the traditional Byzantine-tolerant algorithms, in the manner of a decentralized approach. In addition, we theoretically analyze and prove the convergence of CMFL under different election and selection strategies, which coincides with the experimental results. 
\end{abstract}

}

\maketitle

\IEEEdisplaynontitleabstractindextext

%
\IEEEpeerreviewmaketitle

\IEEEraisesectionheading{\section{Introduction}\label{sec:introduction}}

%
%
%
%

\IEEEPARstart{N}{owadays}, data arise from a wide range of sources, such as mobile devices, commercial, industrial and medical activities. These data are further used for training the Artificial Intelligence (AI) models applied in a variety of fields. The conventional AI methods always require uploading the source data to a central server. However, this is usually impractical due to data privacy or commercial competition. Federated Learning (FL) \cite{konecny2017federated}, which allows multiple devices to train a shared global model without uploading the local source data to a central server, is an effective way to solve the aforementioned problem. In FL settings, multiple clients (also known as participants) are responsible for model training and uploading the local gradients, while the central server is responsible for the model aggregation. A single round of FL mainly follows the following four steps: (1) the multiple clients download a global model from the server, and train their local models on their local datasets; (2) the clients upload the local gradients to the server, and the server aggregates the received multiple local gradients to construct the global gradient; (3) the server uses the global gradient to update the global model; (4) the clients download the global model to the local to continue the next training round. The above operations will be repeated until the algorithm converges. 


It is fascinating that FL can perform model training without uploading source data, McMahan \etal \cite{McMahanMRHA17} proposed that FL can achieve a similar test accuracy as the centralized method based on the full training dataset while providing stronger privacy guarantees. However, Lyu \etal \cite{DBLP:journals/corr/abs-2003-02133} proposed that the conventional FL is vulnerable to malicious attacks from the Byzantine clients and the central server. For example, the Byzantine clients upload false gradients to affect the performance of the global model, which may lead to training failure \cite{Blanchard2017Byzantine}\cite{247652}. Besides, the presence of malicious server is widely considered in FL\cite{8737416}\cite{9293091}\cite{DBLP:journals/corr/BonawitzIKMMPRS16}\cite{DBLP:journals/corr/abs-2109-14236}, and Hu \etal \cite{abs-2010-10996} proposed that external attacks on the central server will cause the entire learning process to terminate. In recent years, there have been a lot of works to solve the security problem of FL. Some works aim to design a robust aggregation rule to reduce the negative impact of malicious gradients\cite{munozgonzalez2019byzantinerobust}. For example, Blanchard \etal \cite{Blanchard2017Byzantine} proposed a Byzantine-tolerant algorithm Krum, which can tolerate Byzantine workers via aggregation rule with resilience property. Similarly, Yin \etal \cite{pmlr-v80-yin18a} proposed two robust distributed gradient descent algorithms based on coordinate-wise median and trimmed mean operations for Byzantine-robust distributed learning. Chen \etal \cite{10.1145/3154503} proposed a simple variant of the classical gradient descent method based on the geometric median of means of the gradients, which can tolerate the Byzantine failures. Some other works detect Byzantine attackers and remove the malicious local gradients from aggregation by clustering\cite{9054676}\cite{yadav2021clustering}. Different from the aforementioned works, Li \etal \cite{li2020learning} propose a framework that trains a detection model on the server to detect and remove the malicious local gradients. Although they can guarantee the convergence and Byzantine-tolerant, they cannot provide effective defense in the presence of malicious servers\cite{kairouz2021advances}.  


As for the further comment, the typical FL requires a central server to complete the gradient aggregation procedure, thus it is difficult to find a fully trusted server in actual scenarios. Beyond that, the entire FL system will be paralyzed if the server suffers a malicious attack. Therefore, a lot of works are devoted to designing a serverless FL framework to reduce the risk of single point failure\cite{pappas2021ipls}. Some existing works design serverless FL frameworks by learning from network protocols such as P2P \cite{roy2019braintorrent}\cite{lalitha2019peertopeer} and Gossip\cite{hu2019decentralized}\cite{10.1007/978-3-030-22496-7_5}. These approaches treat clients as network nodes, which communicate with each other according to the improved network protocol and complete the local training and the aggregation of the global model. Besides, other approaches employ blockchain technology to complete the work of the server to achieve a serverless FL framework\cite{8733825}\cite{9019859}\cite{8843900}\cite{8905038}. They treat clients as blockchain nodes, record the local gradients uploaded on the block, and then make the leading who completes the PoW (Proof of Work) to ensure the aggregation procedure. However, few works present the theoretical convergence analysis of the serverless FL, leading to the lack of performance guarantee.


In this paper, we comprehensively consider Byzantine attacks of both clients and the central server, and design a serverless \underline{F}ederated \underline{L}earning framework under \underline{C}ommittee \underline{M}echanism (CMFL), in which some participating clients are appointed as committee members, and the committee members take the responsibility for monitoring the entire FL training process and ensuring the dependable aggregation of the local gradients. The CMFL consists of three components: the scoring system, the election strategy and the selection strategy. The scoring system plays a role in distinguishing different clients. The election strategy is responsible for selecting some clients who can represent the majority of clients to become members of the new committee. Based on different considerations, we propose two opposite selection strategies to make the committee members verify local gradients uploaded by clients. A selection strategy is designed to ensure the robustness of the training process, where the committee members accept those local gradients who are similar to their own local gradients but reject those local gradients who are significantly different from their own ones. The other selection strategy is designed to accelerate the convergence of the global model in a non-attack scenario, where the committee members accept those local gradients who are different from their own local gradients but reject those gradients who are similar to their own ones. {Compared to some existing Byzantine-tolerant algorithms based on local gradient validation, such as Median\cite{pmlr-v80-yin18a}, Trimmed Mean\cite{pmlr-v80-yin18a} and Krum\cite{Blanchard2017Byzantine}, CMFL achieves better performance and higher robustness over various datasets, illustrated by the experimental results. Besides, CMFL can achieve a higher security level due to its decentralized setting.} Theoretical analysis on the convergence of the model is further presented for the performance guarantee, which illustrates the impact of the proposed election and selection strategies. Extensive experiments further demonstrate the outperformance of CMFL compared with both the typical and Byzantine-tolerant FL models, coinciding with the theoretical analysis on the efficiency of the proposed election and selection strategies. In summary, we highlight the contributions as follow:

\begin{itemize}
	\item We propose a serverless FL framework: CMFL, with the ability to monitor the gradient aggregation procedure and prevent both malicious clients and the server from hindering the training of the global model. 
  \item We propose an election strategy and two selection strategies suitable for different scenarios, which can ensure the robustness of the algorithm or accelerate the training process.
	\item We give the proof and analysis of the convergence of the proposed serverless FL framework for the theoretical guarantee, which considers the influence of election and selection strategies on the performance of the global model. 
	\item We conduct extensive experimental results to show that CMFL has a faster model convergence rate and better model performance than the typical and Byzantine-tolerant FL models.
\end{itemize}

The remainder of this paper is organized as follows. Section \ref{relatedWork} surveys related work. Section \ref{background} briefly outlines the background and problem formulation of FL. Section \ref{CMFL} introduces our proposed framework. We show the convergence analysis in Section \ref{convergenceAnalysis} and complexity analysis in Section \ref{ComplexityAnalysis}. Experimental results and analysis are summarized in Section \ref{experiments}. We conclude the paper in Section \ref{conclusion}. Finally, Supplement gives the convergence proof.




\section{Related Work}
\label{relatedWork}

Konečný \etal \cite{konecny2015federated} introduce a new distributed optimization setting in machine learning in 2015. Based on this setting the concept of FL is proposed\cite{konecny2017federated}, which aims to train an efficient centralized model in a scenario where the training data is distributed across a large number of clients. However, these frameworks suffer from heterogeneous clients, failing to achieve satisfactory performance on the Non-IID dataset. To further handle such issues, several works extend FL to Non-IID dataset. Li \etal \cite{li2020convergence} presents the theoretical guarantees under Non-IID settings and analyzes the convergence of FedAvg. Li \etal \cite{li2020federated} propose a framework \textit{FedProx} with convergence guarantees to tackle heterogeneity on the Non-IID dataset. Yue \etal \cite{zhao2018federated} proposed a strategy to mitigate the negative impact of Non-IID data by share a small subset of data between all the clients. Briggs \etal \cite{9207469} improve the pe{}rformance of FL on the Non-IID dataset by introducing a hierarchical clustering step to separate clusters of clients. {For the study on data heterogeneity, Haddadpour \etal \cite{DBLP:journals/corr/abs-1910-14425} generalize the local stochastic and full gradient descent with a new scheme, periodic averaging, to solve nonconvex optimization problems in FL. Dinh \etal \cite{DBLP:journals/ton/DinhTNHBZG21} proposed FEDL, a FL algorithm which can handle heterogeneous user data without any assumptions except strongly convex and smooth loss functions. Liu \etal \cite{DBLP:journals/tpds/LiuCCZ20} proposed momentum federated learning (MFL) to accelerate the convergence, and they establish global convergence properties of MFL and originate an upper bound on the convergence rate.} Different from the aforementioned works, the proposed framework provides a new perspective to enhance the performance on the Non-IID setting.

The attractiveness of Federated learning relies on the trainable centralized model on user equipment without uploading user data. However, such a framework is vulnerable to Byzantine attackers due to the lack of identity authentication for local gradients. Lots of works have designed a series of Byzantine-tolerant algorithms to further ensure the robustness of the training process. For example, Blanchard \etal \cite{Blanchard2017Byzantine} proposed \textit{Krum}, which aims to select the global model update based on the Euclidean distance between the local models. Yin \etal \cite{pmlr-v80-yin18a} proposed \textit{Median} and \textit{Trimmed Mean}, which are designed to remove extreme local gradients to ensure the robustness of the algorithm. The Median method constructs a global gradient, where each element is the median of the elements in the local gradients with the same coordinate, while the Trimmed Mean method removes the maximum and minimum fraction of elements in the local gradients, and then performs a weighted average on the remaining local gradients to construct the global gradient. Muñoz-González \etal \cite{munozgonzalez2019byzantinerobust} propose a Hidden Markov Model to learn the quality of local gradients and design a robust aggregation rule, which discards the bad local gradients in the aggregation procedure. The aforementioned algorithms are all trying to ensure the robustness of FL by designing a more appropriate aggregation mechanism. However, these works can not provide effective defense in the presence of malicious servers. 

In real applications, commercial competition makes it difficult to find a fully trusted central server among the participants. In addition, server error or malicious server will also cause irreparable damage to the FL system. In this way, many serverless FL frameworks are proposed to solve these problems. Among these approaches, some of them achieve a serverless FL framework by imitating existing network protocols. For example, Abhijit \etal \cite{roy2019braintorrent} proposed a P2P serverless FL framework, where any two clients exchange information end-to-end and update their local models at each epoch. Hu \etal \cite{hu2019decentralized} used the Gossip protocol to complete the model aggregation process, which takes on the role of the central server. Besides, other works build a serverless FL framework based on the blockchain system. For example, Kim \etal \cite{8733825} proposed a blockchain FL architecture, in which they divide the blockchain nodes into devices and miners. The device nodes provide data, train the model locally, and upload the local gradients to their associated miner in the blockchain network. Miner nodes exchange and verify all the local gradients. Although these works have obtained the corresponding performance to some degree, they lack the theoretical analysis of the model convergence under serverless FL framework and consideration of Byzantine attacks of clients. 


To deal with the limitations of the existing works, we proposed a serverless FL framework under committee mechanism. In scenarios that consider the Byzantine attacks, the framework can be efficient Byzantine-tolerant of malicious clients and servers. In scenarios without considering the Byzantine attacks, the framework can achieve better performance of the global model on the Non-IID dataset. Beyond that, we present the theoretical analysis of the convergence under the framework.




\section{Background}
\label{background}

\subsection{Federated Learning}

A typical FL framework consists of a central server and multiple clients. The server maintains a global model, and each client maintains a local model. At the beginning of training, the global model and all local models will be initialized randomly. And then the following steps will be performed at each communication round to continue the training process\cite{konecny2017federated}:

\begin{enumerate}[1.]
\item The server randomly selects a subset of clients, which then download the global model to the local.
\item Each client in the subset performs a certain number of Stochastic Gradient Descent (SGD)\cite{stich2019local}\cite{wang2019adaptive} and computes the local gradient.
\item The clients in the subset send their local gradients to the server.
\item The server receives the local gradients and performs the Federated Averaging (FedAvg) algorithm \cite{pmlr-v54-mcmahan17a} to construct a global gradient, which is used to update the global model. 
\end{enumerate}  

The above steps will be iterated until the algorithm converges or the model accuracy meets the requirements. 

\subsection{Problem Formulation}

We consider the typical FL setup with total $K$ clients. The $k$-th client for $k = 1,...,K$ owns a local dataset $\D_k$, which contains $n_k=|\D_k|$ data samples. We denote the user-defined loss function for sample $\xi$ and model parameter vector $\w$ as $f(\w,\xi)$, the local objective function  $F_k(\w)$ on the $k$-th client can be written as follows:

\begin{equation}
F_k(\w) = \frac{1}{n_k}\sum_{\xi \in \D_k}f(\w,\xi).
\end{equation}We consider the following global objective function:

\begin{equation}
F(\w) = \sum_{k=1}^K p_kF_k(\w),
\end{equation}where $p_k = n_k/\sum_{k=1}^K n_k$ denotes the weight of the dataset on the $k$-th client. Formally, $\nabla F_k(\w_{k,i}^t)$ denotes the local gradient over dataset $\mathcal{D}_k$. Assume that at round $t$ the $k$-th client trains its local model $\w_{k,i}^{t}$ over mini-batch $\B_{k,i}^{t}$ for $i$ iterations of SGD, where $\B_{k,i}^{t}$ is randomly sampled from $\D_k$. Then the $k$-th client computes the local gradient $g_k(\w_{k,i}^{t},\B_{k,i}^{t})$ by the following formula:

\begin{equation}
g_k(\w_{k,i}^{t},\B_{k,i}^{t}) = \frac{1}{|\B^t_{k,i}|}\sum_{\xi \in \B_{k,i}^{t}}\nabla f(\w_{k,i}^{t},\xi).
\end{equation}The $g_k(\w_{k,i}^{t},\B_{k,i}^{t})$ is used to update the local model $\w_{k,i}^{t}$ as follows:

\begin{equation}\label{localUpdate2}
\w_{k,i+1}^{t} = \w_{k,i}^{t} - \eta_{i}^t g_k(\w_{k,i}^{t},\B_{k,i}^{t}),
\end{equation}where $\eta_{i}^t$ represents the local learning rate at iteration $i$ of round $t$ and $\tau$ represents the maximal local SGD iterations. And after $\tau$ iterations of local updating, the local gradient $g_{k}(\w_{k,\tau}^{t},\B_{k,\tau}^{t})$ is sent to the server to construct a global gradient as follows:

\begin{equation}
\label{aggregation}
\overline{g}^{t} = \sum_{k \in S^{t}}p_{k,S^t} g_k(\w_{k,\tau}^{t},\B_{k,\tau}^{t}), 
\end{equation}where $S^{t}$ denotes the subset of clients and $p_{k,S^t} = n_k/\sum_{k\in S^t}n_k$ is the weight of the dataset on the $k$-th client of $S^t$. The $\overline{\w}^{t}$ is updated at each round as follows:

\begin{equation}
\label{globalUpdate}
\overline{\w}^{t+1} = \overline{\w}^{t} - \eta_t \overline{g}^{t},
\end{equation}where $\eta_t$ represents the global learning rate.

\section{Committee Mechanism based Federated Learning}
\label{CMFL}

The typical FL system is vulnerable to Byzantine attacks and malicious servers due to its disability to implement a serverless framework and the lack of verification for the uploaded local gradients. The key insight of CMFL is to utilize the committee mechanism to implement a decentralized framework. Under such a decentralized framework, we appoint some training clients as the committee members, which are responsible for filtering the local gradients uploaded by the remaining clients. The committee members must reach a consensus on which the uploaded gradient should be used for the global updating. In this way, the aggregation process is controlled by all the committee members rather than one untrustworthy central server. As long as the number of the honest committee members is greater than that of the malicious members, attacks from malicious clients become meaningless, since the decision of the committee depends on the majority of the committee members. In order to guarantee the honesty of committee members and achieve a secure aggregation, we design a novel committee mechanism, including a scoring system, selection strategy, election strategy, and committee consensus protocol. The detailed introduction of the proposed framework and the training progress is involved in this section and the further theoretical analysis of the framework is illustrated in Section \ref{convergenceAnalysis}.


\subsection{Framework of CMFL}\label{FrameworkOfCMFL}

\begin{figure*}
	\centering
	\includegraphics[width=17cm]{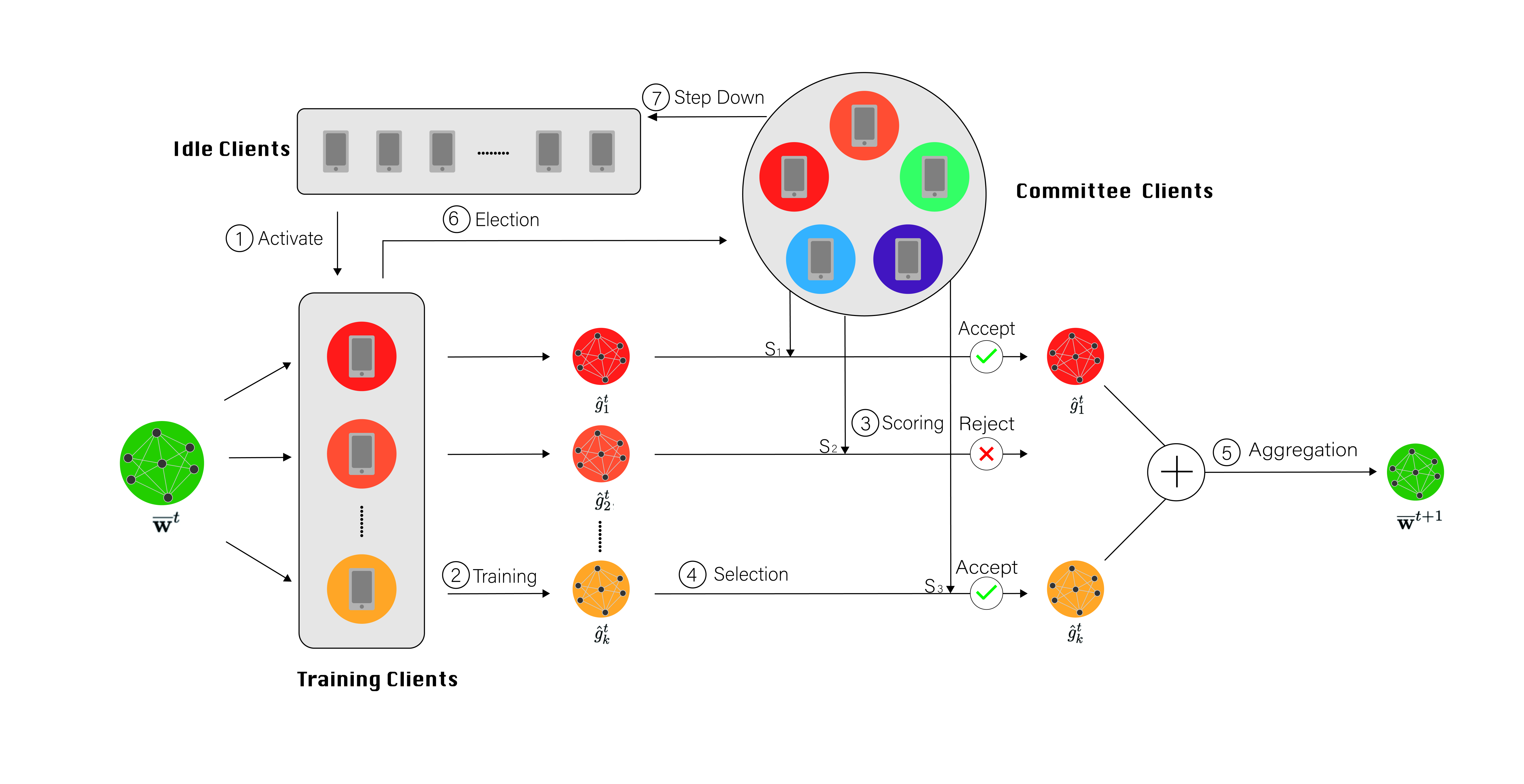}
	\caption{The training process of CMFL are as follows: (1) Training clients are selected randomly from the idle clients; (2) The training clients and committee clients download the global model $\overline{\w}^{t}$ and start training, then send their local gradients $\hat{g}_k^{t}$ to the committee clients; (3) The committee client assigns a score to each client according to the scoring system; (4) The committee selects the aggregation clients according to the selection strategy; (5) The local gradients uploaded by the aggregation clients are used to construct the global gradient $\overline{\w}^{t+1}$; (6) New committee clients are elected from the training clients according to the election strategy; (7) The committee clients of the last round are added to the idle client list.}
	\label{fig:CMFL}
\end{figure*}

In the CMFL framework shown in Figure \ref{fig:CMFL}, the clients are divided into three categories: \textit{training client}, \textit{committee client}, and \textit{idle client}. At each round, the following steps are performed to complete the training process:

\begin{itemize}
\item \textbf{Activate.} A part of the idle clients are activated to be the training clients, which participate in the training at this round.
\item \textbf{Training.} The training clients and the committee clients download the global model to the local for training, while the idle clients stay idle until the next round. The training clients and the committee clients perform SGD over their local dataset and compute the local gradients. The difference is that the local gradients on the training clients are used to update the global model, while the local gradients on the committee clients are used to verify the gradients uploaded by the training clients. 
\item \textbf{Scoring.} The training clients send their local gradients to each committee client and the committee clients assign a score on them according to an established scoring system. 
\item \textbf{Selection.} Only the qualified gradients according to the set selection strategy will be used to construct the global gradient. 
\item \textbf{Aggregation.} The clients who meet the selection strategy are responsible for completing the aggregation procedure, which is called the \textit{aggregation client}. 
\item \textbf{Election.} An election strategy is designed to complete the replacement of committee members. A part of the training clients who meet the election strategy become the new committee clients.
\item \textbf{Step Down.} The prior committee clients become the idle clients, waiting for the next participate.
\end{itemize}

In a decentralized framework, we design a committee consensus protocol based on Practical Byzantine Fault Tolerance(pBFT)\cite{PBFT}, to complete the Selection, Aggregation, and Election process.

Supposed that $C$ clients will be selected as committee clients, the local model of $c$-th committee client for $c = 1,...,C$ at round $t$ is expressed as $\w_{c,\tau}^{t}$. Assume that we have a total of $K$ clients, and $m$ gradients will be accepted at each round, which are verified by $C$ committee members. Also, we represent the committee client set, the training client set and the aggregation client set at round $t$ as $S_c^{t}$, $S_b^{t}$ and $S_a^{t}$ respectively. The relevant symbols are shown in Table \ref{notation}.


\begin{table}[h]
\caption{Notation}
\label{notation}
\centering
\begin{tabular}{c c}
\toprule
  Notation & Description \\
  \midrule
  $K$ & Number of total clients \\
  $T$ & Number of maximal communication rounds \\ 
  $\tau$ & Number of maximal SGD iterations\\
  $\D_k$ & Local dataset\\
  $\B_{k,i}^{t}$ & Mini-batch randomly sampled from $\D_k$\\
  $\w_{k,i}^{t}$ & Local model parameter\\
  $\overline{\w}^{t}$ & Global model parameter\\
  $\w_k^*$ & Optimal local model parameter\\
  $\w^*$ & Optimal global model parameter\\
  $\w_{c,i}^{t}$ & Local model parameter of committee client\\
  $f(\w,\xi)$ & Overall loss function\\
  $F_k(\w)$ & Local objective function\\
  $g_k(\w_{k,i}^{t},\B_{k,i}^{t})$ & Mini-batch local gradient for $i$ iteration\\
  $\nabla F_k(\w_{k,i}^{t})$ & Full local gradient for $i$ iteration\\
  $\hat{g}_k^t$ & Local gradient for $\tau$ iterations\\
  $F^*_k$ & Optimal value of the local objective function\\
  $F^*$ & Optimal value of the global objective function\\
  $S_a^{t}$ & Aggregation client set\\
  $S_b^{t}$ & Training client set\\
  $S_c^{t}$ & Committee client set\\
  $m$ & Number of clients in $S_a^{t}$\\
  $n$ & Number of clients in $S_b^{t}$\\
  $C$ & Number of clients in $S_c^{t}$\\
  $\p_k^c$ & Score of the $k$-th client assigned by the $c$-th client\\
  $\p_k$ & Final score of the $k$-th client\\
\bottomrule
\end{tabular}

\end{table}

\subsection{Committee Mechanism}

A committee system is set up to complete the screening of local gradients. The committee system consists of the scoring system, the election strategy, the selection strategy and the committee consensus protocol. 

\subsubsection{Scoring System}

The key insight of the scoring system is to distinguish between the honest and malicious gradients by calculating their Euclidean distance. The Euclidean distance of two honest gradients is lower than that of an honest gradient and a malicious gradient. Based on this insight, the scoring system is designed as comparing the Euclidean distance of two gradients, where the clients who upload the honest gradients can obtain higher scores while the clients who upload the malicious gradients will obtain the lower score. Assume that the local gradient on the $k$-th training client at round $t$ is denoted as $\hat{g}_{k}^{t} = g_{k}(\w_{k,\tau}^{t},\B_{k,\tau}^{t})$, and the local gradient on the the $c$-th committee client at round $t$ is expressed as $\hat{g}_c^{t} = g_c(\w_{c,\tau}^{t},\B_{k,\tau}^{t})$. The score $\p_k^c$ of the $k$-th training client assigned by the $c$-th committee client is computed as follows:

\begin{equation}
\p_k^c = \frac{1}{||\hat{g}_k^{t} - \hat{g}_c^{t}||^2_2}.
\end{equation}
Since $\hat{g}_k^t$ and $\hat{g}_c^t$ are local gradients generated from different clients, we assume that $\hat{g}_k^t \neq \hat{g}_c^t$ for any $k \in S_b^t, c \in S_c^t$. We define 

\begin{equation}\label{computeScore}
\begin{split}
&\p_k =\frac{1}{\frac{1}{C}\sum_c^C ||\hat{g}_k^{t} - \hat{g}_c^{t}||^2_2} = \frac{C}{\sum_c^C\frac{1}{\p_k^c}}
\end{split}
\end{equation} as the final score of the $k$-th training client. The scoring principle is mainly based on the Euclidean distance between the local gradients $\hat{g}_k^{t}$ and $\hat{g}_c^{t}$. Another insight remains that Byzantine attackers usually replace some local gradients with malicious gradients, which directly increases the Euclidean distance between these gradients and the honest gradients. As a result, when the proportion of malicious clients is within a tolerable range, the score of the malicious training clients is expected to be lower than the honest training clients. However, in a scenario without Byzantine attacks, the score represents the degree of heterogeneity of clients. A higher score means a higher degree of heterogeneity.

\subsubsection{Selection Strategy}
\label{selection}

We design the selection strategy for determining which uploaded gradient is used to update the global model. Based on our scoring system, we can achieve a secure aggregation by accepting the gradients with high scores, since the malicious gradients obtain low scores. Besides, the convergence analysis and experimental result show that the opposite strategy performs better in a non-attack scenario. Therefore, two opposite selection strategies are designed for different considerations as follows:  

\begin{itemize}
\item \textbf{Selection Strategy \uppercase\expandafter{\romannumeral1}.} The selection strategy \uppercase\expandafter{\romannumeral1} selects several local gradients with relatively higher scores to construct a global gradient for the update of the global model. In other words, we hope that the local gradients which are similar to the committee gradients in the Euclidean space will participate in the construction of the global gradient. In practice, we sort the local gradients according to their scores and only accept the top $\alpha\%$ of them. We design the selection strategy for the consideration of robustness. Those malicious gradients and honest gradients are far from each other in the Euclidean space, and choosing the gradients close to the committee gradients under the condition that the committee clients are honest can achieve a more robust aggregation process. However, it is difficult for those clients with obvious differences to be selected to participate in the aggregation procedure, which makes it hard for the global model to learn the comprehensive data characteristic, resulting in a decline in performance. Overall, selection strategy \uppercase\expandafter{\romannumeral1} is suitable for FL scenarios with Byzantine attacks. 
\item \textbf{Selection Strategy \uppercase\expandafter{\romannumeral2}.} The selection strategy \uppercase\expandafter{\romannumeral2} selects several local gradients with relatively lower scores to construct a global gradient for the update of the global model. That is, we hope that the local gradients which are different from the committee gradients in the Euclidean space will participate in the construction of the global gradient. Similar to the selection strategy \uppercase\expandafter{\romannumeral1}, we sort the local gradients according to their scores and only accept the bottom $\alpha\%$ of them in practice. We design the selection strategy \uppercase\expandafter{\romannumeral2} for the consideration of convergence rate and performance of the global model. Constructing a global gradient by selecting those local gradients with obvious differences allows the global model to learn more comprehensively, gaining a faster convergence and better performance of the global model. However, this strategy can not provide robustness guarantees, mainly since that the malicious gradients will be preferentially used in global learning, leading to a sharp drop in the performance of the global model. Overall, selection strategy \uppercase\expandafter{\romannumeral2} is suitable for FL scenarios without Byzantine attacks. 
\end{itemize}

\subsubsection{Election Strategy}
\label{election}

The election strategy is designed for guaranteeing the honesty of committee clients. The committee members reach a consensus on their decisions, and the committee's decision depends on the majority of the committee members. However, the malicious clients mixed into the committee may interfere with the committee's decision-making. Therefore, the committee must guarantee that its honest members are more than malicious members. Otherwise, the committee can not filter out the malicious gradients. We sort the training clients according to their scores and then select these training clients closed to the middle position as the committee clients for the next round. We design the election mechanism based on the following two considerations:

\begin{itemize}
  \item \textit{Robustness}. According to the above analysis, as malicious training clients will get lower scores than honest training clients, they are expected to locate at the end of the sorted sequence. Therefore, choosing the training client closed to the middle or upper position prevents the malicious training clients from becoming the committee clients in the next round, thereby guaranteeing the security of the global model. Although there may still be a small number of malicious clients mixed into the committee members when the proportion of malicious clients is relatively large, it is difficult to affect the judgment of the whole committee because of their exceeding small proportion of the committee members.
  \item \textit{System Stability}. Those training clients with the highest scores are not chosen to be the new committee members in order to avoid the system from relying too much on the initialization of committee members. When we choose those training clients with the highest scores to form a new committee, the learning direction of the global model will be completely determined by the initial committee members. It is because those clients with a large Euclidean distance between the local gradient and the committee gradients are not only difficult to be selected to participate in the aggregation procedure, but also lost the opportunity to run for the next round of committee members. This is in line with our intuition that “committee members should be those who can represent the majority”.
\end{itemize}

\subsubsection{Committee Consensus Protocol}
\label{CommitteeConsensus}

In our decentralized framework, the committee clients must reach a consensus to complete the scoring, aggregation, selection, and election process. Thus we design a committee consensus protocol, which is inspired by the pBFT\cite{8069090}. The designed committee consensus protocol(CCP) is as follows(at round $t$):

\begin{enumerate}
\item After the scoring process, each committee client obtains the scores of the training clients. Then every committee client broadcast its scores to the other committee clients. Each committee client is able to calculate the total score of every training client by Eq. \eqref{computeScore}.
\item A committee client $p$ is selected as the primary committee client by random, while other committee clients are regarded as the replicate committee clients. 
\item Each committee client decides its aggregation set $S_{a,c}^t$ according to the selection strategy. Since all scores are broadcasted among committee clients, the aggregation sets among honest committee clients are the same. 
\item The primary committee client creates a request $\langle$ \textit{Request, $S_{a,p}^t$, operation, timestamp} $\rangle$ to ask whether its $S_{a,p}^t$ is correct. Then the primary committee client broadcasts the request to all the replicate committee clients. The operation in the request is to aggregate the local gradients uploaded by aggregation clients as Eq.\eqref{aggregation}
\item All the replicate committee clients execute the request. Each of them checks whether the $S_{a,p}^t$ is the same as its own $S_{a,c}^t$. If so, it performs the aggregation process as Eq. \eqref{aggregation}. After aggregation, it checks whether the result is consistent with the request. If so, the executing result $\langle$ \textit{Reply, timestamp, $S_{a,p}^t$, response} $\rangle$ is returned to the primary committee client.
\item The primary committee client checks whether it has received at least $\lfloor C/2 \rfloor + 1$ identical results from replicate committee clients. If so, the consensus is reached; otherwise, the primary committee client should be reassigned and steps 3)-6) should be repeated.
\item Similarly, steps 3)-6) are repeated to reach consensus on new committee client set $S_c^{t+1}$(However, the aggregation client set $S_a^t$ in step 3)-6) should be replaced by new committee client set $S_c^{t+1}$).
\item The primary committee client broadcasts the global model to all other clients. The next training round is ready to start.  
\end{enumerate}

\begin{algorithm}
\caption{Committee Consensus Protocol}
\begin{algorithmic}[1]
\label{algorithm_CCP}
\REQUIRE $t$, $S_c^t$, $S_b^t$\\ 
\ENSURE global model $\w^t$, new committee client set $S_c^{t+1}$\\ 
\FOR{$c \in S_c^t$}
\FOR{$k \in S_b^t$}
\STATE The $c$-th committee client broadcasts score $\p_k^c$ to other committee clients.
\ENDFOR
\ENDFOR  
\FOR{$c \in S_c^t$}
\FOR{$k\in S_b^t$}
\STATE The $c$-th committee client calculates the total score $\p_k$ of the $k$-th training client.
\ENDFOR
\ENDFOR
\STATE {The committee performs voting($S_a^t$, aggregtion) as Algorithm \ref{voting} and gets the global model $\w^t$. }
\STATE The committee performs voting($S_c^{t+1}$, None) as Algorithm \ref{voting}.
\end{algorithmic}
\end{algorithm}
\begin{algorithm}
\caption{The voting algorithm}
\begin{algorithmic}[1]
\label{voting}
\REQUIRE The client set $S$ and the opertation $o$\\ 
\ENSURE {The result $r$ of the operation $o$}\\ 
\WHILE{No consensus on $S$}
\STATE The primary committee client $p$ is selected from committee clients by random.
\FOR{$k \in S_c^t$}
\STATE The $k$-th committee client decides its client set $S^k$ according to the corresponding strategy (selection strategy/election strategy). 
\ENDFOR
\STATE {$p$ performs the operation $o$ and gets $r$.}
\STATE $p$ creates a request $\langle$ \textit{Request, $S^p$, $o$, timestamp} $\rangle$ and broadcasts it to all the replicate committee clients.
\FOR{$k \in S_c^t \setminus \{p\}$}
\STATE The $k$-th replicate committee client receives the request from $p$.
\IF{$S^k == S^p$}
\STATE The $k$-th committee client performs the operation $o$.
\STATE Return $\langle$ \textit{Reply, timestamp, $S^p$, response} $\rangle$ to $p$.
\ENDIF
\IF{$p$ receives at least $\lfloor C/2\rfloor + 1$ response}
\STATE Consensus is reached on $S$.
\ENDIF 
\ENDFOR
\STATE { Return $r$.}
\ENDWHILE
\end{algorithmic}
\end{algorithm}

\subsection{Training Algorithm}

In this section, we introduce the serverless training algorithm of CMFL. Firstly, all clients randomly initialize the local model and some clients are randomly selected as the committee clients. In each communication round, the algorithm performs the following five steps:
\begin{enumerate}
  \item \textit{Random Sampling}: A part of the clients from the non-committee clients will be selected randomly as the training clients while the other clients become the idle clients. At round $t$ all training clients and committee clients download the global model from the primary committee client as the local model.
  \item \textit{Local Training}: All training clients and committee clients perform $\tau$ rounds of SGD over the local datasets and compute the local gradients. The training clients send their local gradients to each committee client for verification. 
  \item \textit{Gradient Verification}: Each committee client assigns a score on each training client according to the scoring system. Then they execute CCP to reach a consensus on aggregation client set $S_a^t$. 
  \item \textit{Global Model Updating}: In CCP, the local gradients uploaded by clients in $S_a^t$ are aggregated for constructing the global gradient, which is used to update the global model according to the Eq. \eqref{globalUpdate} when the consensus is reached.

  \item \textit{Election of New Committee Members}: The committee clients execute CCP to reach a consensus on new committee clients set $S_c^{t+1}$ 
\end{enumerate}

The algorithm repeats the above five steps until the algorithm converges or $t$ exceeds the defined maximum communication round $T$.

\begin{algorithm}
\caption{The training algorithm of CMFL}
\begin{algorithmic}[1]
\label{algorithm_CMFL}
\REQUIRE $\tau$, $T$, $K$, $m$, $C$, $\eta_t$, $\eta_i^t$\\ 
\ENSURE target global model $\w^T$\\ 
\STATE Initialize $\w^{1}$, $S_c^{1}$ and $S_b^{1}$ randomly.
\FOR{$t=1$ to $T$}
\FOR{$k \in S_b^{t} \cup S_c^{t}$}
\FOR{$i=1$ to $\tau$}
\STATE The $k$-th client runs the SGD over the local dataset by $\w_{k,i}^{t} \Leftarrow \w_{k,i-1}^{t} - \eta_i^t g_{k}(\w_{k,i-1}^{t},\B_{k,i-1}^{t})$.
\ENDFOR
\ENDFOR
\FOR{$k \in S_b^{t}$}
\FOR{$c \in S_c^{t}$}
\STATE The $k$-th training client send its local gradient $\hat{g}_k^{t}$ to the $c$-th committee client.
\STATE The $c$-th committee client assign a score $\p_k^c$ on the $k$-th training client/local gradient according to the scoring system.
\ENDFOR
\ENDFOR
\STATE The committee clients execute CCP to complete the selection, aggregation, and election process.
\STATE The $S_b^{t}$ be reinitialized to form the $S_b^{t+1}$.
\FOR{$k\in S_b^{t+1} \cup S_c^{t+1}$}
\STATE The $k$-th client downloads the global model from the primary committee client.
\ENDFOR
\ENDFOR
\end{algorithmic}
\end{algorithm}

\section{Theoretical Analysis}

In this section, we show the theoretical analysis of CMFL, including convergence analysis and complexity analysis. 

\subsection{Convergence Analysis}
\label{convergenceAnalysis}

In this section, we conduct the convergence analysis of the proposed framework CMFL. First, we introduce some basic assumptions used for the convergence analysis in Section \ref{assumption}. Second, we introduce two definitions that facilitate our analysis in Section \ref{convergence_definition}. Finally, we present our result and analysis for convergence of CMFL in Section \ref{convergence_result}. The proof of the convergence is presented in the Supplement.

For the purpose of convergence proof and analysis, we define $F^* = \min_{\w}F(\w) = F(\w^*)$ as the optimal value of the global objective function, where $\w^*$ denotes the optimal global model. In the same way we define $F_k^* = \min_{\w}F_k(\w) = F_k(\w_k^*)$ as the optimal value of the $k$-th client's local objective function, where $\w_k^*$ denotes the optimal local model of $k$-th client.

\subsubsection{Assumptions}\label{assumption}

First, we introduce the assumptions used for convergence analysis.

\begin{assumption}\label{A1} 
(Lipschitz Gradient). \textit{$F_1,...,F_K$ are all $\mathcal{L}$-smooh: for all $\mathbf{v}, \w \in R^n$, $k = 1, ..., K$, $F_k(\mathbf{v}) \leq F_k(\w) + (\mathbf{v}-\w)^T\nabla F_k(\w) + \frac{L}{2}||\mathbf{v}-\w||^2_2$.}
\end{assumption}

\begin{assumption}\label{A2} 
($\mu$-strongly Convex Gradient). \textit{$F_1,...,F_K$ are all $\mu$-strongly convex: for all $\mathbf{v}, \w \in R^n$, $k = 1, ..., K$, $F_k(\mathbf{v}) \geq F_k(\w) + (\mathbf{v}-\w)^T\nabla F_k(\w) + \frac{\mu}{2}||\mathbf{v}-\w||^2_2$}
\end{assumption}

\begin{assumption}\label{A3} 
(Bounded Variance). \textit{For the mini-batch $B_{k,i}^t$ uniformly sampled randomly from $k$-th client's dataset $\D_k$, the resulting stochastic gradient is unbiased: $E[g_k(\w_{k,i}^t,\B_{k,i}^t)]=\nabla F_k(\w_{k,i}^t)$ for all $k=1,...,K, t=1,...,T, i=1,...,\tau$. And the variance of stochastic grandient in each client is bounded: $E||g_k(\w_{k,i}^t,\B_{k,i}^t)-\nabla F_k(\w_{k,i}^t)||^2 \leq \sigma^2$.}
\end{assumption}

\begin{assumption}\label{A4}
(Bounded Gradient). \textit{The local gradient's expected squared norm is uniformly bounded: $\mathbb{E}||g_k(\w_{k,i}^t,\B_{k,i}^t)||^2 \leq G^2$ for all $k = 1,...,K,t=1,...,T, i=1,...,\tau$.}
\end{assumption}

\begin{assumption}\label{A5}
(Bounded Objective Function). \textit{For any aggregation client set $S_a\notin \varnothing$ and the optimal committee client set $S_c^* \notin \varnothing$, the difference of local optimal objective function is bounded: $\mathbb{E}[||\sum_{k \in S_a}p_{k,S_a}F_k^*-\sum_{k' \in S_c^*}p_{k',S_c^*}F_{k'}^*||] \leq \kappa^2$, where $S_c^*$ satisfies that $S_c^* = \arg\min_{S_c}\sum_{k\in S_c}p_{k,S_c}F_k^*$ and $|S_c^*| = C$.} 
\end{assumption}



Assumption \ref{A1} and \ref{A2} are standard conditions in FL setting\cite{9252927}\cite{9261995}\cite{9003425}\cite{amiri2020convergence} and many common machine learning optimization algorithms meet these assumptions, such as the $\ell_2$-norm regularized linear regression, logistic regression, and softmax classifier \cite{li2020convergence}. Assumption \ref{A3} is a form of bounded variance between the local objective functions and the global objective function\cite{8664630}, and Assumption \ref{A4} is fairly standard in nonconvex optimization literature\cite{pmlr-v48-reddi16}\cite{pmlr-v97-ward19a}\cite{9069945}\cite{9148862}. They are widely used in the FL convergence analysis, such as Li \etal \cite{li2020convergence} and Cho \etal \cite{cho2020client}. Assumption \ref{A5} is used to constrain the optimal objective function deviation between the aggregation client set and the committee client set caused by the committee mechanism. For the need of convergence proof, we define $S_c^*$ as the set which contains $C$ clients with the smallest optimal local objective function $F_k^*$. 

\subsubsection{Definition}\label{convergence_definition}

We introduce two related definitions for the convenience of analysis as follows.

\begin{definition}\label{heterogeneity}
\textbf{(Degree of Heterogeneity).} \textit{We use 
\begin{equation}
\Gamma = F^* - \sum_{k=1}^Kp_kF_k^* = \sum_{k=1}^Kp_k(F_k(\w^*)-F_k(\w_k^*))
\end{equation}
to quantify the degree of heterogeneity among the clients.}
\end{definition}
Li \etal \cite{li2020convergence} proposed this definition, which is widely used in the convergence analysis of FL on Non-IID dataset\cite{9337227}. In the Non-IID FL scenarios, the $\Gamma$ remains nonzero and its value reflects the heterogeneity of the data distribution. In the IID FL scenarios, with the growth of $K$ the $\Gamma$ gradually goes to zero.

\begin{definition}\label{CommitteeSkew}
\textbf{(Aggregation-Committee Gap).} \textit{For any aggregation client set $S_a^t$, we define  
\begin{equation}
\begin{split}
&\varphi(S_a^{t},\w) \\
=& \frac{\mathbb{E}[\sum_{k \in S^{t}_a}p_{k,S_a^t}F_k(\w)-\sum_{k' \in S_c^*}p_{k',S_c^*}F_{k'}^*]}{F(\w) - \sum_{k=1}^Kp_kF_k^*} \geq 0,
\end{split}
\end{equation}
where $\mathbb{E}$ denotes the expectation over all randomness in the previous iterations, and $S_c^*$ denotes the optimal committee client set.}
\end{definition}
$\varphi(S_a^t,\w)$ changes with the changes of $S_a^t$ and $\w$ during training. An upper bound $\varphi_{max}$ and a lower bound $\varphi_{min}$ are defined to obtain a conservative error bound in the convergence analysis:
\begin{equation}
\begin{split}
&\varphi_{min} = \min_{S_a^t,\w}\varphi(S_a^t,\w),\varphi_{max} = \max_{S_a^t}\varphi(S_a^t,\w^*).
\end{split}
\end{equation}

\subsubsection{Convergence Result and Analysis}\label{convergence_result}

We analyze the convergence of Algorithm \ref{algorithm_CMFL} in this section and find an upper bound of $\mathbb{E}[F(\overline{\w}^{(T)})] - F^*$, which denotes the convergence error of the global model after $T$ rounds:

\begin{theorem}\label{theorem1}
(Convergence of Committee Mechanism based Federated Learning). \textit{Under Assumption \ref{A1} to \ref{A5}, Definition \ref{heterogeneity} to \ref{CommitteeSkew} and the learning rate $\eta_t$, where $ \eta_t = \frac{1}{\mu(t+\gamma)}$ and $\gamma = \frac{4L}{\mu}$, the error after $T$ rounds of CMFL satisfies}
\begin{small}
\begin{equation}
\begin{aligned}
&\mathbb{E}[F(\overline{\w}^{T})] - F^*\\
\leq& \frac{1}{T+\gamma}\left[ \frac{4L(32\tau^2G^2 + \sum_{k=1}^Kp_k\sigma_k^2) + 24L^2\kappa^2}{3\mu^2 \varphi_{min}} \right. \\
&\left. + \frac{8L^2\Gamma}{\mu^2}+\frac{L\gamma||\overline{\w}^{1} - \w^*||^2}{2}\right] + \frac{8L\Gamma}{3\mu}\left(\frac{\varphi_{max}}{\varphi_{min}}-1\right),
\end{aligned}
\end{equation}
\end{small}
\end{theorem}
where $L$, $\mu$ and $\sigma$ represent the constant light indicated in the Assumption \ref{A1}, \ref{A2} and \ref{A3}. $G$ and $\kappa$ represent the upper bound values defined in Assumption \ref{A4} and \ref{A5} .$\Gamma$ denotes the heterogeneity among the clients according to the Definition \ref{heterogeneity}. These are all constant while $\varphi_{min}$ and $\varphi_{max}$ will be different depending on the selection strategy. The impact of selection strategy on $\varphi_{min}$ is analyzed below. CMFL affects the performance of the global model by altering $\varphi_{min}$. The proof of the theorem is shown in Supplement. 

\textbf{The impact of selection strategy on $\varphi_{min}$.} According to the definition, the value of $\varphi$ is positively correlated with $\sum_{k\in S_a^t}p_{k,S_a^t}F_k(\w)$, which represents the average local loss of the model over the aggregation client set $S_a^t$. Under our framework, the lower bound of $\varphi(S_a^{t},\overline{\w}^{t})$ defined as $\varphi_{min}$ affects the convergence rate of the global model, where $\overline{\w}^{t}$ represents the global model at round $t$. In general, those clients whose data set distribution is similar to that of the majority have a relatively low local loss, while others have a relatively high local loss. We call the former \textit{universal clients} and the latter \textit{extreme clients}. The tendency to choose a universal client for aggregation results in low $\varphi_{min}$, and the tendency to choose an extreme client for aggregation results in high $\varphi_{min}$. The building of the election strategy ensures that committee members can represent the majority, and the scoring system is designed as the clients similar to committee members can get higher scores, so clients with high scores are more likely to have a low local loss. When we adopt selection strategy \uppercase\expandafter{\romannumeral1}, we get a low $\varphi_{min}$. Instead, when we adopt selection strategy \uppercase\expandafter{\romannumeral2}, we get a high $\varphi_{min}$.

\textbf{Effect of $\varphi_{min}$ on convergence rate.} Note that a higher $\varphi_{min}$ results in faster convergence at the rate $\mathcal{O}(\frac{1}{T\varphi_{min}})$. That is, adopting the selection strategy \uppercase\expandafter{\romannumeral1} make the convergence of the global model slows down, while the selection strategy \uppercase\expandafter{\romannumeral2} accelerates the convergence of the global model, which is verified in the following experiments. However, considering the reality of Byzantine attacks, the first selection strategy is a more appropriate choice.

\subsection{Complexity Analysis}
\label{ComplexityAnalysis}
In this section, we analyze the computation complexity and communication complexity of the proposed framework CMFL. For each complexity analysis, the time and overhead of CMFL are considered. {Recall the previous notations, we define $C$ as the number of committee clients, $n$ as the training clients.} Noted that $m$ represents the number of aggregation clients. 

\subsubsection{Computaion Complexity}

There are five phases related to computation complexity as follows:

\begin{itemize}
  \item \textbf{Local Training}. Each training client and committee client performs SGD locally before they upload their local gradients. The computation overhead of $k$-th client is $\mathcal{O}(|\D_k|\cdot|\w|)$, where $|\D_k|$ is the number of data samples, $|\w|$ is the model size. Assumed that $|\D|$ represents the average value of $|\D_k|$ over all clients, the total computation overhead is {$\mathcal{O}((n + C)|\D|\cdot|\w|)$}. Because each client performs local training in parellel, the computation time of local training is $\mathcal{O}(|\D|\cdot|\w|)$. 
  \item \textbf{Scoring}. In the phase of scoring, each committee client assigns a score to each training client. According to the scoring system, the computation overhead of scoring for one committee client is {$\mathcal{O}(n \cdot |\w|^2)$}. The total computation overhead of scoring is {$\mathcal{O}(n\cdot C \cdot |\w|^2)$}. As each committee client performs scoring operation in parallel, the computation time of scoring is {$\mathcal{O}(n \cdot |\w|^2)$}. 
  \item \textbf{Aggregation}. In the phase of aggregation, only $m$ local gradients are used for aggregation, so the computation overhead of aggregation for one committee client is $\mathcal{O}(m\cdot |\w|)$. According to the CCP, each committee client performs aggregation so the total computation overhead of aggregation is {$\mathcal{O}(m\cdot C\cdot|\w|)$}. Since the aggregation process is performed in parallel, the computation time of aggregation is $\mathcal{O}(m\cdot |\w|)$. 
  \item \textbf{Selection}. In the phase of selection, each committee client sorts the received local gradients and selects $m$ local gradients for aggregation. The computation overhead is {$\mathcal{O}(nlogn)$}, which can be ignored since it is much smaller than the computation overhead of the above three phases. 
  \item \textbf{Election}. In the phase of the election, each committee client determines its new committee client set based on the sorted local gradients. The computation overhead is $\mathcal{O}(1)$, which can be ignored because it is much smaller than the computation overhead of the above four phases. 
\end{itemize}

\subsubsection{Communication Complexity}
Assumed that each client owns the same maximum bandwidth, and $r$ represents the max transmission rate, the communication time of transferring data $s$ is computed as $T_{transmission} = s/r$. There are three phases related to communication complexity as follows:
\begin{itemize}
  \item \textbf{Gradient Uploading}. After local training each training client uploads its local gradient to all the committee clients, so the communication overhead of uploading for one client is {$\mathcal{O}(C \cdot |\w|)$}. The total communication overhead of uploading is {$\mathcal{O}(n \cdot C \cdot |\w|)$}. Since each training client performs the uploading process in parellel, the communication time of uploading is {$\mathcal{O}(C \cdot |\w|/r)$}. 
  \item \textbf{Global Model Downloading}. After global aggregation, the primary committee client broadcasts the global model to all training clients in the next round. The total communication overhead of downloading is {$\mathcal{O}(n \cdot |\w|)$}, thus the communication time of downloading is {$\mathcal{O}(n \cdot |\w|/r)$}. 
  \item \textbf{Broadcasting}. Committee clients should perform CCP to reach consensus, in this phase the primary committee client broadcast $S_a^t$ and $S_c^{t+1}$ to other committee clients, which occur communication overhead. However, this communication overhead can be ignored because it is much smaller than the communication overhead of uploading and downloading. 
\end{itemize}


{Besides, system heterogeneity is a widespread problem in the field of Federated Learning. The differences in the computational performance of clients lead to various time consumptions, which becomes the bottleneck limiting the efficiency of the Federated Learning system. Some clients even drop out during the training process. There are some methods to alleviate the performance degradation caused by system heterogeneity, such as setting a maximum waiting time or a minimum number of received gradients. Indeed, system heterogeneity is an important issue in Federated Learning, and more works need to be carried out to address it.}

\section{Experiments}
\label{experiments}

In this section, we first present our experimental setup in Section \ref{setup}, which includes the datasets, models, and experimental environment. Then, we evaluate our proposed framework CMFL by five sets of following experiments, and the results and analysis are presented in Section \ref{nomaltrainingexperiment} to Section \ref{CommitteeAnalysisEvaluation}.

\begin{enumerate}
  \item \textbf{Normal Training Experiment.} We test the performance of CMFL without considering the Byzantine attack and compare it with typical FL.
  \item \textbf{Robustness Comparative Experiment.} We evaluate the Byzantine resilience of CMFL and compare it with several Byzantine-tolerant algorithms. 
  \item \textbf{Hyperparameter Analysis Experiment.} We vary the hyper-parameter $\alpha$, $\omega$, and $\epsilon$ and show how it affects the performance.
  \item \textbf{Efficiency Experiment.} We evaluate the efficiency of CMFL and compare it with other decentralized FL frameworks while considering the computation and communication overhead.
  \item \textbf{Committee Member Analysis Experiment.} We track the malicious clients that have been mixed into the committee members to analyze the impact of these malicious clients on the performance of the global model. 
\end{enumerate}

\subsection{Experimental Setup} \label{setup}

\subsubsection{Datasets and Models}

{We evaluate CMFL over three datasets with a Non-IID setting, including FEMNIST, Sentiment140, and Shakespeare.}

\begin{itemize}
  \item \textbf{FEMNIST-AlexNet.}\cite{caldas2019leaf}  FEMNIST (Federated Extended MNIST) is a real-world distributed dataset formed by a specific division of the EMNIST dataset. This dataset is used to train a model for handwritten digit/character recognition tasks, which contains 805263 images of $28 \times 28$ pixels, divided into 64 categories, each category represents a type of handwritten digit/character ($0-9$ and $a-z$). The FEMNIST dataset divides the EMNIST dataset into 3550 parts in a specific way and stores them on each client to simulate a real federated learning scenario. We use the convolutional neural network AlexNet as the basic experimental model for this image classification dataset.
  \item \textbf{Sentiment140-LSTM.} \cite{sent140} Sentiment140 is a real-world distributed dataset which focus on the text sentiment analysis task, including 1,600,000 tweets extracted using the Twitter API. The data in this dataset has been annotated (0 = negative, 2 = neutral, 4 = positive). It can be used to discover the sentiment of a brand, product, or topic on Twitter. We regard each Twitter account as a client and choose a two-layer LSTM binary classifier as the basic experimental model. The LSTM binary classifier includes 256 hidden units with pretrained 300D GloVe embedding\cite{pennington2014glove}.  
  \item {\textbf{Shakespeare-LSTM.} \cite{DBLP:conf/aistats/McMahanMRHA17} Shakespeare is a real-world distributed dataset built from \textit{The Complete Works of William Shakespeare}. Each speaking role corresponds to a different device, and we choose a two-layer LSTM classifier as the basic experimental model, which contains 100 hidden units with an 8D embedding layer. There are 80 classes of characters in dataset. The model takes a sequence of 80 characters as input, embeds each character into a learned 8-dimensional space, and outputs one character for each training sample after 2 LSTM layers and a densely connected layer.}
\end{itemize}

\begin{table*}[h]
\caption{Statistics of datasets}
\label{noniidlevel}
\centering
\begin{tabular}{c c c c c}
\toprule
  Dataset & Number of devices & Total samples & Mean & Stdev \\
  \midrule
  FEMNIST & 3,550 & 805,263 & 226.83 & 88.94 \\
  Sentiment140 & 772 & 40,783 & 53 & 32\\ 
  {Shakespeare} & {143} & {517,106} & {3616} & {6808}\\
\bottomrule
\end{tabular}

\end{table*}

The Non-IID levels of these two datasets are shown in Table \ref{noniidlevel}.

\subsubsection{Environment}

\hspace{1.0em} \textbf{Global Setup}: At the beginning of each communication round, $10\%$ of the idle clients are activated to participate in the training, while the rest clients wait for the next selection. At one communication round, each client runs one iteration of SGD for local training. Besides, the initial committee clients are selected from all clients at random. In order to better demonstrate the experimental effect, we set different learning rates $\eta$ in different datasets.

\textbf{Machines}: We perform our experiment on a commodity machine with Intel Core CPU i9-9900X containing a clock rate of 3.50 GHz with 10 cores and 2 threads per core. And we utilize Geforce RTX 2080Ti GPUs to accelerate training. The learning model is implemented in Python 3.7.6 and Tensorflow 1.14. 

\textbf{Hyper-parameter Notation}: In each communication round, $10\%$ of the whole clients are activated to participate in the training of that round, $\omega\%$ of which become the committee clients and the rest become the training clients. Each round $\alpha\%$ of the training clients become the aggregation clients, whose local gradients are accepted to use for the constructing of the global gradient. $\epsilon$ presents the percentage of malicious clients.

\subsection{Nomal Training Experiment}\label{nomaltrainingexperiment}

\subsubsection{Experiment Setting}

In this experiment we set $\alpha = 40$, $\omega=40$ and $\epsilon=0$ to simulate a non-attack FL scene. The $\eta$ in FEMNIST and Sentiment140 is $0.001$ and $0.005$ respectively. In this experiment, we compared typical FL and CMFL under these two selection strategies:

\begin{itemize}
	\item \textbf{Typical FL.} Typical FL uses all the local gradients uploaded by the training client for the construction of global gradients.
	\item \textbf{CMFL with Selection Strategy \uppercase\expandafter{\romannumeral1}.} The original intention of this selection strategy is to resist Byzantine attacks. The committee accepts the local gradients uploaded by the high-score training clients while rejects those uploaded by low-score training clients.
	\item \textbf{CMFL with Selection Strategy \uppercase\expandafter{\romannumeral2}.} This selection strategy is opposite to selection strategy \uppercase\expandafter{\romannumeral1}, accepting the local gradients uploaded by the low-score clients while rejects those uploaded by high-score training clients.
\end{itemize}

\begin{figure*}[htbp]
	\centering
	
	\subfigure[FEMNIST-Accuracy]{
		\begin{minipage}[t]{0.3\linewidth}
			\centering
			\includegraphics[width=5.5cm]{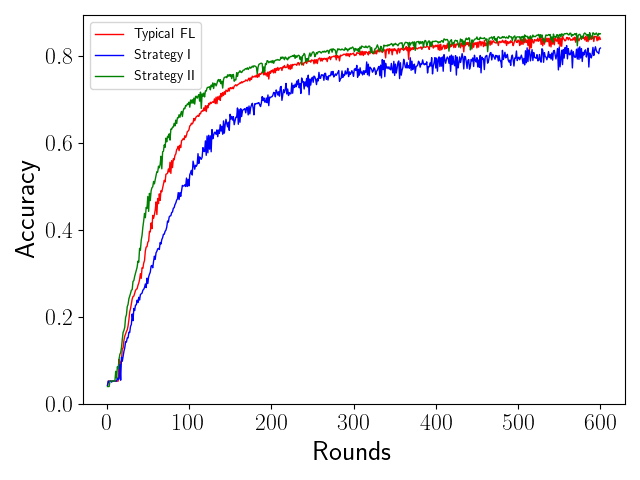}
		\end{minipage}%
	}%
  \subfigure[Sentiment140-Accuracy]{
    \begin{minipage}[t]{0.3\linewidth}
      \centering
      \includegraphics[width=5.5cm]{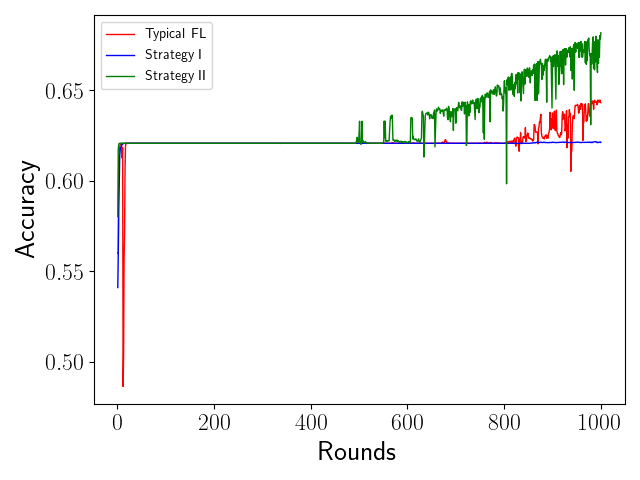}
    \end{minipage}%
  }%
  \subfigure[Shakespeare-Accuracy]{
    \begin{minipage}[t]{0.3\linewidth}
      \centering
      \includegraphics[width=5.5cm]{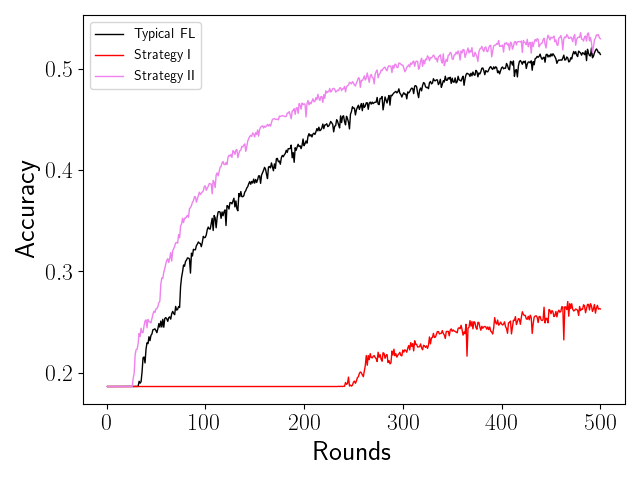}
    \end{minipage}%
  }%

	\subfigure[FEMNIST-Loss]{
		\begin{minipage}[t]{0.3\linewidth}
			\centering
			\includegraphics[width=5.5cm]{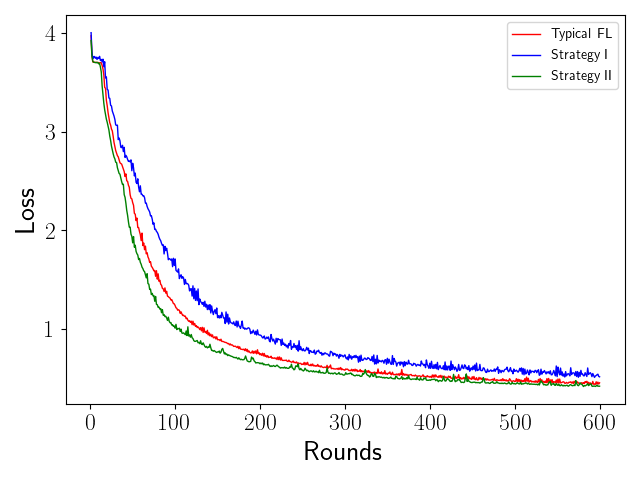}
		\end{minipage}%
	}%
  \subfigure[Sentiment140-Loss]{
    \begin{minipage}[t]{0.3\linewidth}
      \centering
      \includegraphics[width=5.5cm]{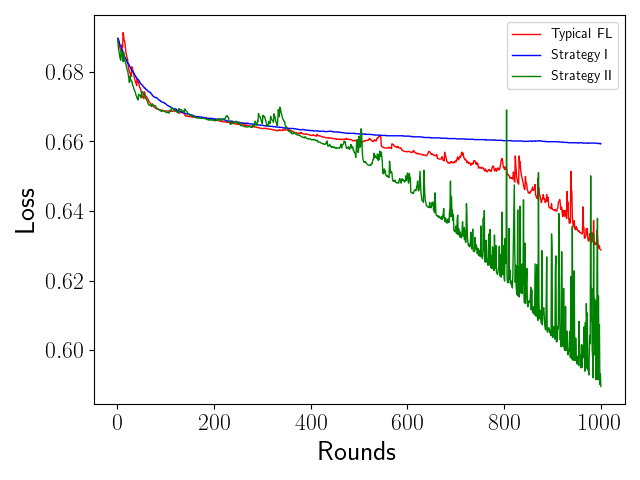}
    \end{minipage}%
  }%
  \subfigure[Shakespeare-Loss]{
    \begin{minipage}[t]{0.3\linewidth}
      \centering
      \includegraphics[width=5.5cm]{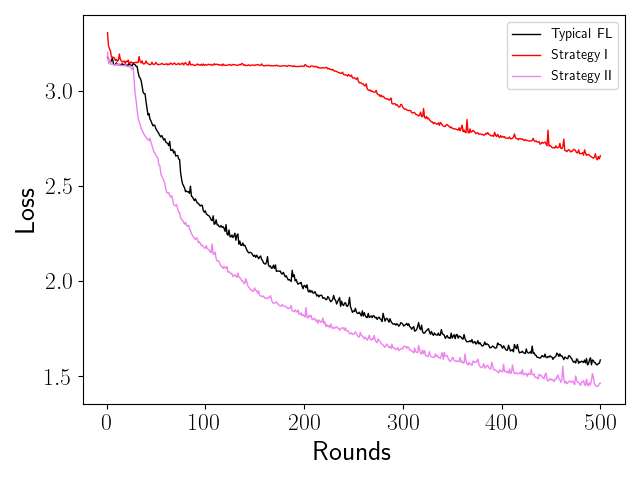}
    \end{minipage}%
  }%
	
	\centering
	\caption{The performance of global model among the typical FL and CMFL with two selection strategies.}
	\label{fig:CMFL_FL_strategy}
\end{figure*}
\subsubsection{Result Analysis}

\begin{figure*}[htbp]
  \centering
  
  \subfigure[]{
    \begin{minipage}[t]{0.3\linewidth}
      \centering
      \includegraphics[width=6cm]{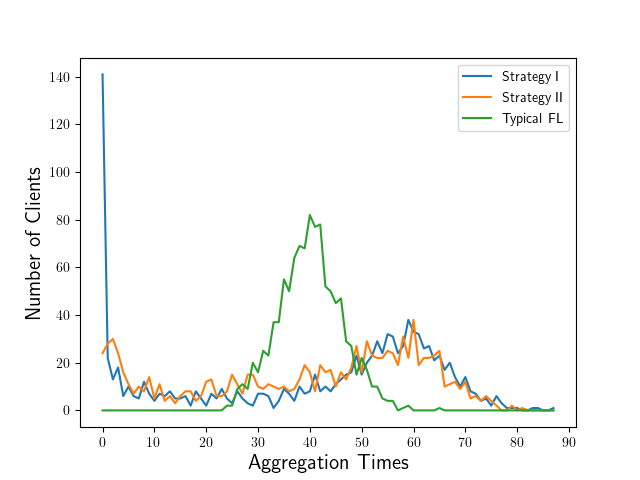}\label{fig:cleint_aggregation}
    \end{minipage}%
  }%
  \subfigure[]{
    \begin{minipage}[t]{0.3\linewidth}
      \centering
      \includegraphics[width=6cm]{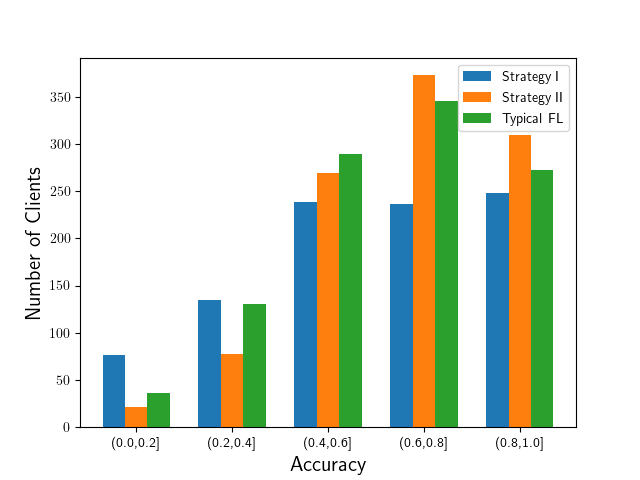}\label{fig:client_acc}
    \end{minipage}%
  }%
  \subfigure[]{
    \begin{minipage}[t]{0.3\linewidth}
      \centering
      \includegraphics[width=6cm]{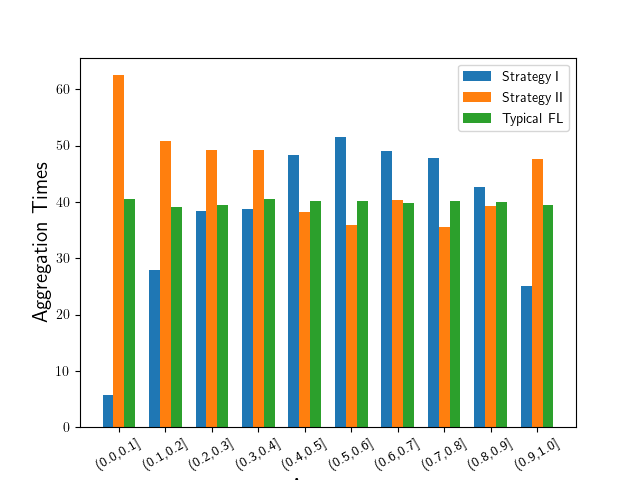}\label{fig:aggregation_acc}
    \end{minipage}%
  }%
  
  \centering
  \caption{Figure (a) shows the number of clients with different aggregation times. Figure (b) shows the number of clients with different accuracy. Figure (b) shows the aggregation times of clients with different accuracy.}
  \label{fig:furtherAnalysis}
\end{figure*}

We show the result in Figure \ref{fig:CMFL_FL_strategy}. Note that the performance of CMFL under selection strategy \uppercase\expandafter{\romannumeral2} is better than that of CMFL under selection strategy \uppercase\expandafter{\romannumeral1} and typical FL. The CMFL under the selection strategy \uppercase\expandafter{\romannumeral1} has the slowest model convergence rate and the worst model accuracy among the three. The result shows that it is a more appropriate design for the committee to accept the local gradients uploaded by the low-score clients when in non-attack FL scene, which can not only enhance the global model performance but also accelerate the convergence of the global model. This is because over the Non-IID dataset, the CMFL under the selection strategy \uppercase\expandafter{\romannumeral1} makes it difficult for a few clients which are quite different from other clients to participate in the aggregation process, resulting in the training of the global model only using part of the data instead of all of the data. {Figure \ref{fig:cleint_aggregation} proves it. We record the aggregation times of each client and plot them as a curve graph. From the curve, we can see that the aggregation times of client under FL conform to a Gaussian distribution, causing FL selects the aggregation randomly. Besides, we found that a high percentage of clients never participate in the aggregation process when performing CMFL under selection strategy \uppercase\expandafter{\romannumeral1}. That is why CMFL under selection strategy \uppercase\expandafter{\romannumeral1} achieves worse performance than typical FL and CMFL under selection strategy \uppercase\expandafter{\romannumeral2}. It is difficult for the global model to learn comprehensive knowledge while using this kind of training algorithm. However, the CMFL under the selection strategy \uppercase\expandafter{\romannumeral2} significantly reduces the number of such clients. In this way more clients can participate in the aggregation process, making the global model learn a more comprehensive knowledge. Figure \ref{fig:client_acc} and \ref{fig:aggregation_acc} explain why CMFL under selection strategy \uppercase\expandafter{\romannumeral2} achieves a better performance than typical FL. We record the testing accuracy of each client after training. Figure \ref{fig:client_acc} shows that CMFL under selection strategy \uppercase\expandafter{\romannumeral2} has least clients with low accuracy, while has most clients with high accuracy. That is because those clients with low accuracy have more opportunities to participate in the aggregation process, which is proved by Figure \ref{fig:aggregation_acc}. Figure \ref{fig:aggregation_acc} shows that the aggregation times of clients with low accuracy in CMFL under selection strategy \uppercase\expandafter{\romannumeral2} are much more than other training methods. By reducing the proportion of clients with low accuracy, CMFL under selection strategy \uppercase\expandafter{\romannumeral2} achieves a better performance than typical FL. } 


Nevertheless, it is not advisable to choose strategy \uppercase\expandafter{\romannumeral2} when considering the Byzantine attack. According to the scoring system we designed, the malicious client will get a lower score than the honest client. Taking the selection strategy \uppercase\expandafter{\romannumeral2} means that Byzantine attackers can easily attack the global model by uploading malicious local gradients. In this scenario, choosing strategy \uppercase\expandafter{\romannumeral1} is a more appropriate choice by avoiding the local gradients with low scores to participate in the aggregation, which is likely to be malicious gradients. Although the effect of CMFL under strategy \uppercase\expandafter{\romannumeral1} in the non-attack FL scenario is slightly weaker than that of the typical FL, it can make the training process more robust in a wider range of realistic scenarios. 

\subsection{Robustness Comparative Experiment}\label{comparativeExperiment}

\subsubsection{Experiment Setting}

In this experiment we set $\alpha = 40$, $\omega=40$ and $\epsilon=10$, and we test the Byzantine resilience of CMFL under two selection strategies by comparing it with the following Byzantine-tolerant algorithms, which are all aim to design a robust gradient aggregation method.

\begin{itemize}
  \item \textbf{Median}\cite{pmlr-v80-yin18a}: This aggregation method first sorts the gradients and selects the median to be the global gradient.
  \item \textbf{Trimmed Mean}\cite{pmlr-v80-yin18a}: This aggregation method first sorts the local gradients, then removes the maximum value of $\beta\%$ and the minimum value of $\beta\%$, and finally aggregates the remaining local gradients to construct the global gradient, where $\beta \in [0,50)$.
  \item \textbf{Krum}\cite{Blanchard2017Byzantine}: This aggregation method assumes that the directions of the local gradients uploaded are relatively similar and sort these local gradients according to their similarity. The original Krum algorithm only selects the first local gradient after sorting as the global gradient.
  \item \textbf{Multi-Krum}\cite{Blanchard2017Byzantine}: The Multi-Krum algorithm is an improved version of the original Krum, which selects the top $\alpha\%$ of the sorted local gradients to construct the global gradient. Compared with the original Krum, Multi-Krum reduces the fluctuation of the global model effect caused by random factors, making the training process more stable. Since Multi-Krum uses more local gradients to construct the global gradients, it can make the global model converge faster than the original Krum.
\end{itemize}

We compare these Byzantine-tolerant algorithms with CMFL considering three different types of attacks: gradient scaling attack, same-value gaussian attack, and back-gradient attack, where the attackers are all aim to compromise some clients and upload the malicious gradients.

\begin{itemize}
	\item \textbf{Gradient Scaling Attack}: The malicious clients multiply each element in the local gradient by a random value $\lambda\in [a, 1)$, where $a$ is a defined constant which indicates the magnitude of the attack. In this experiment we set $a=0.5$. 
	\item \textbf{Same-value Attack}\cite{DBLP:conf/aaai/LiXCGL19}: The malicious clients replace the local gradient with a vector of the same size whose elements are all 0.
	\item \textbf{Back-gradient Attack}\cite{10.1145/3128572.3140451}: The malicious clients replace the local gradient with a vector of the same size in the opposite direction.
\end{itemize}

\subsubsection{Result Analysis}

\begin{figure*}[htbp]
	\centering
	
	\subfigure[Gradient Scaling Attack Loss]{
		\begin{minipage}[t]{0.3\linewidth}
			\centering
			\includegraphics[width=5.5cm]{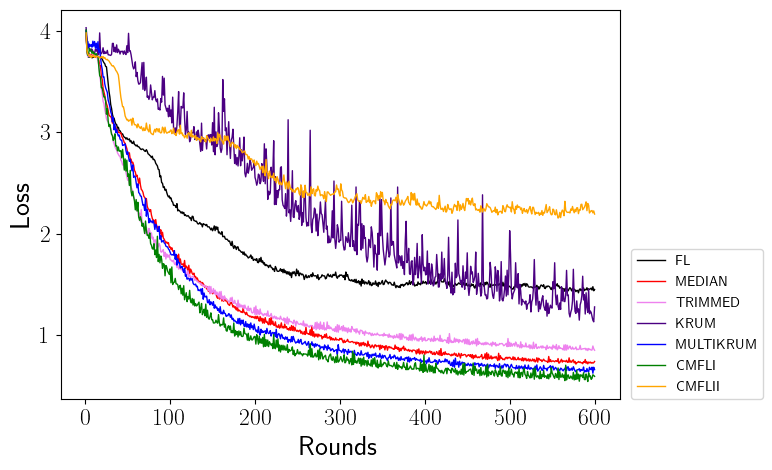}
		\end{minipage}%
	}%
	\subfigure[Same-value Attack Loss]{
		\begin{minipage}[t]{0.3\linewidth}
			\centering
			\includegraphics[width=5.5cm]{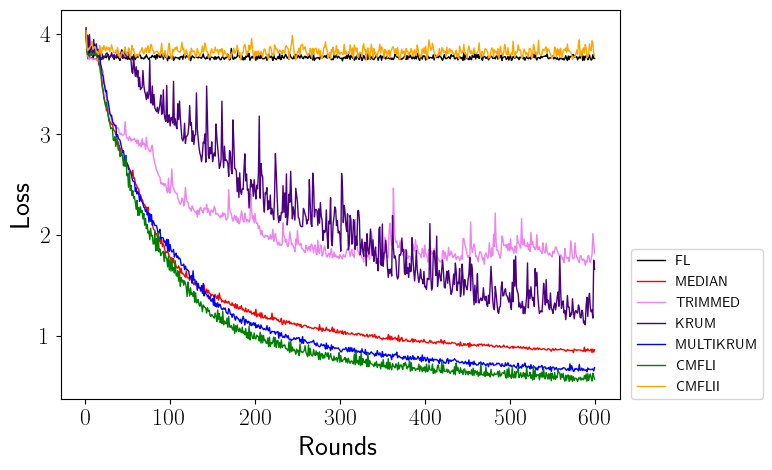}
		\end{minipage}%
	}%
	\subfigure[Back-gradient Attack Loss]{
		\begin{minipage}[t]{0.3\linewidth}
			\centering
			\includegraphics[width=5.5cm]{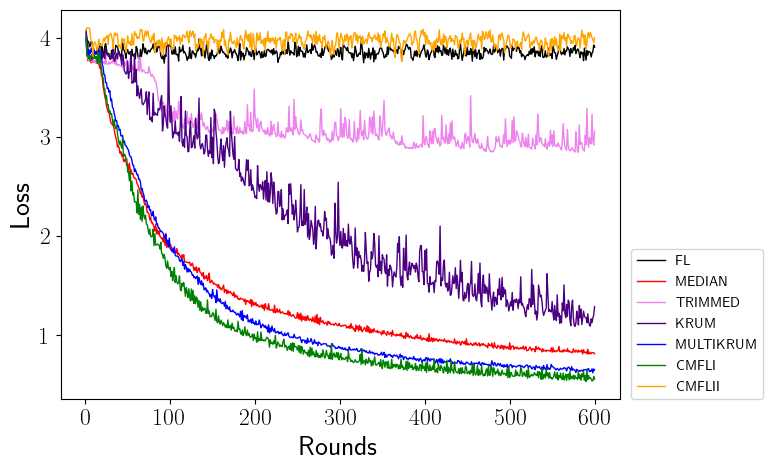}
		\end{minipage}%
	}%
	
	\subfigure[Gradient Scaling Attack Accuracy]{
		\begin{minipage}[t]{0.295\linewidth}
			\centering
			\includegraphics[width=5.5cm]{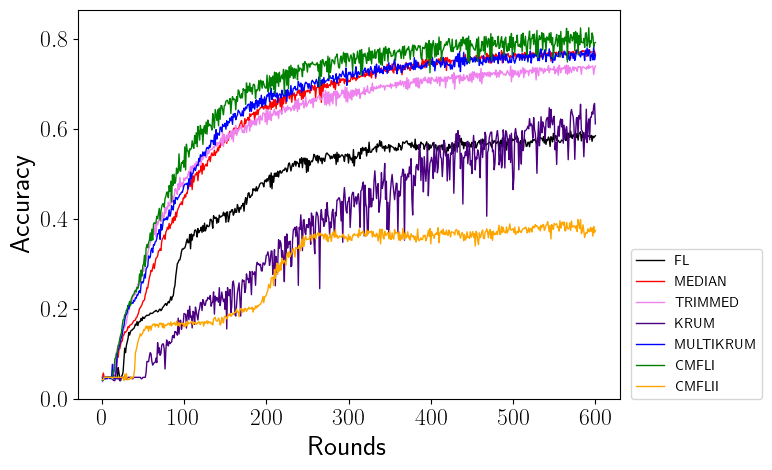}
		\end{minipage}
	}%
	\subfigure[Same-value Attack Accuracy]{
		\begin{minipage}[t]{0.295\linewidth}
			\centering
			\includegraphics[width=5.5cm]{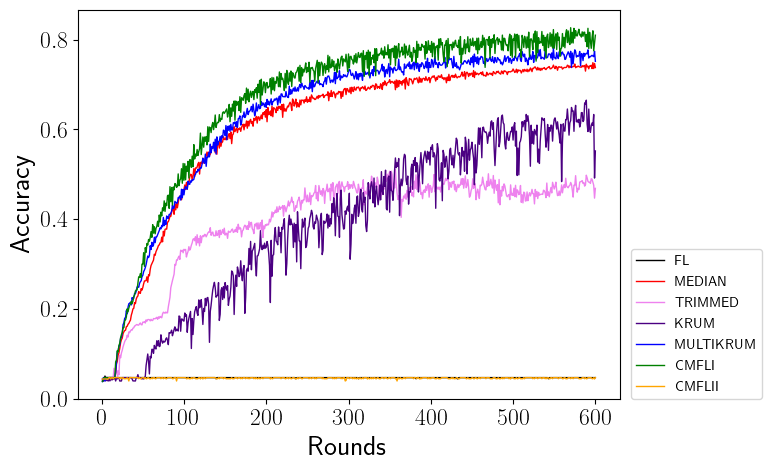}
		\end{minipage}
	}%
	\subfigure[Back-gradient Attack Accuracy]{
		\begin{minipage}[t]{0.295\linewidth}
			\centering
			\includegraphics[width=5.5cm]{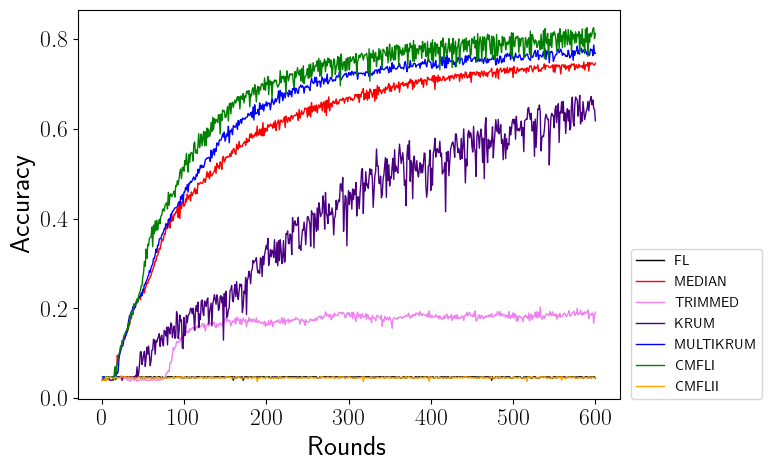}
		\end{minipage}%
	}%
	
	\centering
	\caption{Performance of CMFL compared with other Byzantine-tolerant algorithms under various attacks over FEMNIST dataset. CMFL\uppercase\expandafter{\romannumeral1} represents CMFL under selection strategy \uppercase\expandafter{\romannumeral1} and CMFL \uppercase\expandafter{\romannumeral2} represents CMFL under selection strategy\uppercase\expandafter{\romannumeral2}.}
	\label{fig:EXP1_FEMNIST}
\end{figure*}

\begin{figure*}[htbp]
  \centering
  
  \subfigure[Gradient Scaling Attack Loss]{
    \begin{minipage}[t]{0.3\linewidth}
      \centering
      \includegraphics[width=5.5cm]{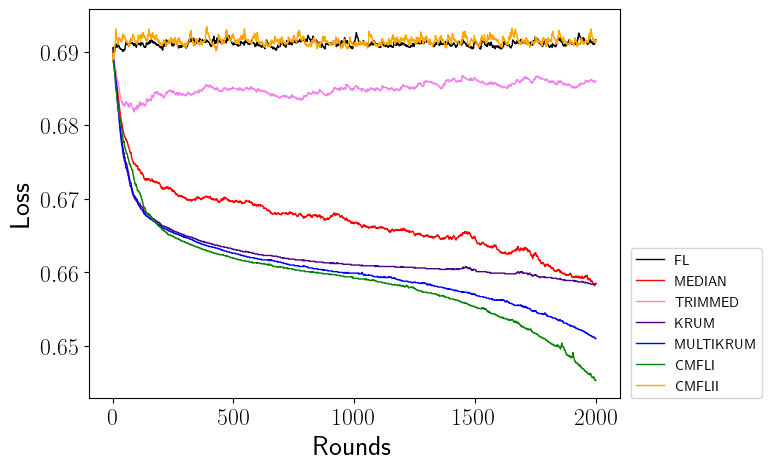}
    \end{minipage}%
  }%
  \subfigure[Same-value Attack Loss]{
    \begin{minipage}[t]{0.3\linewidth}
      \centering
      \includegraphics[width=5.5cm]{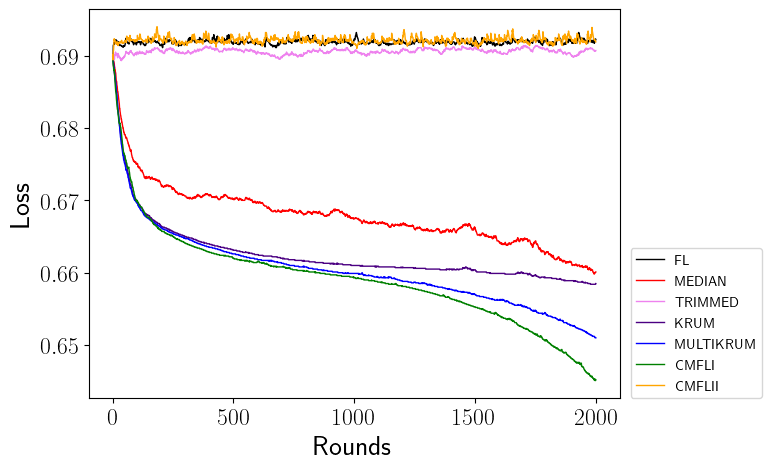}
    \end{minipage}%
  }%
  \subfigure[Back-gradient Attack Loss]{
    \begin{minipage}[t]{0.3\linewidth}
      \centering
      \includegraphics[width=5.5cm]{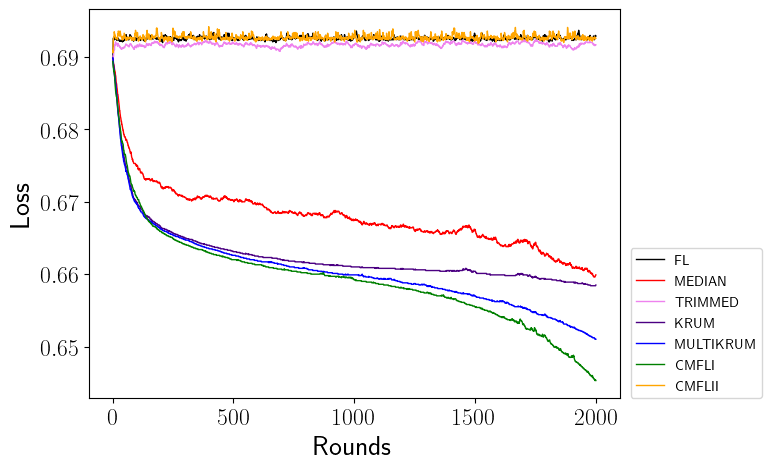}
    \end{minipage}%
  }%
  
  \subfigure[Gradient Scaling Attack Accuracy]{
    \begin{minipage}[t]{0.295\linewidth}
      \centering
      \includegraphics[width=5.5cm]{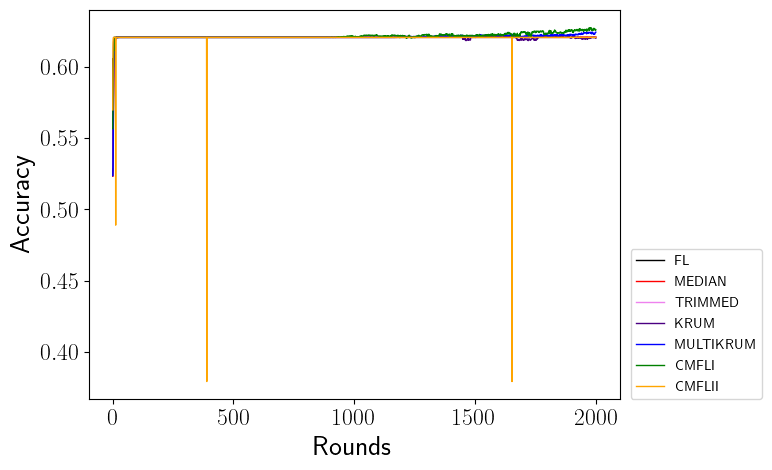}
    \end{minipage}
  }%
  \subfigure[Same-value Attack Accuracy]{
    \begin{minipage}[t]{0.295\linewidth}
      \centering
      \includegraphics[width=5.5cm]{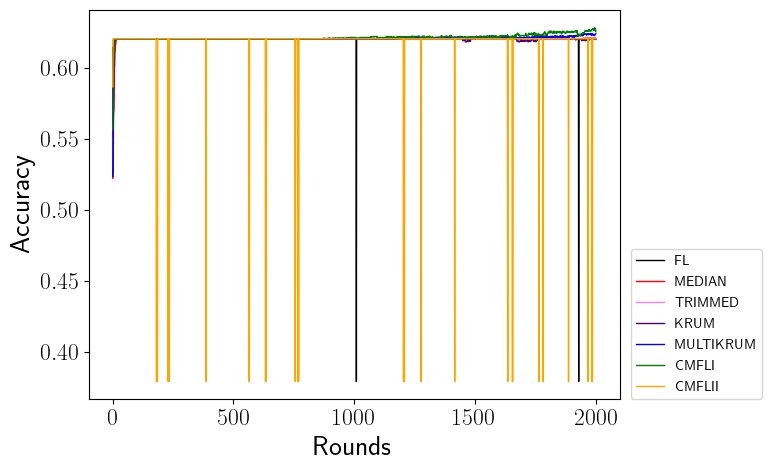}
    \end{minipage}
  }%
  \subfigure[Back-gradient Attack Accuracy]{
    \begin{minipage}[t]{0.295\linewidth}
      \centering
      \includegraphics[width=5.5cm]{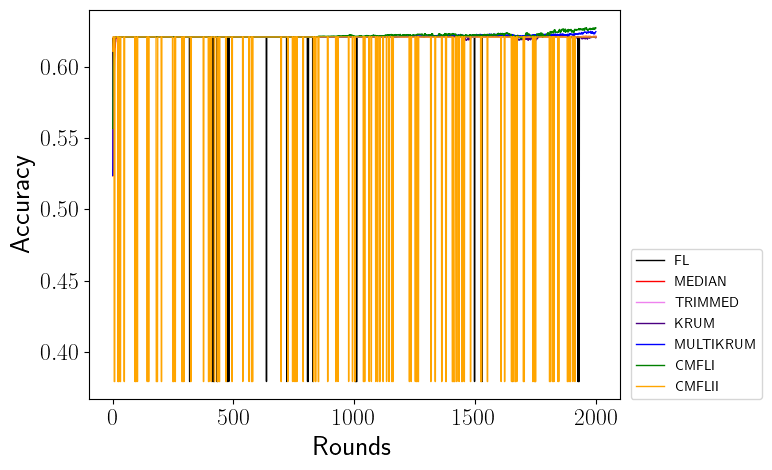}
    \end{minipage}%
  }%
  
  \centering
  \caption{Performance of CMFL compared with other Byzantine-tolerant algorithms under various attacks over Sentiment140 dataset.}
  \label{fig:EXP1_Sent140}
\end{figure*}

\begin{figure*}[htbp]
  \centering
  
  \subfigure[Gradient Scaling Attack Loss]{
    \begin{minipage}[t]{0.3\linewidth}
      \centering
      \includegraphics[width=5.5cm]{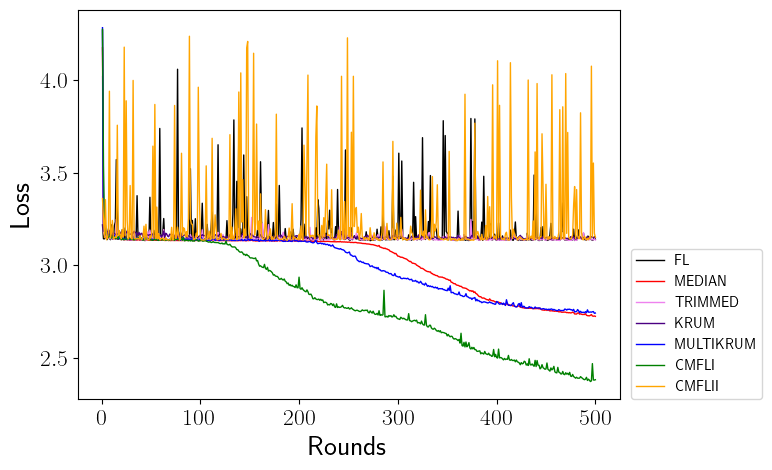}
    \end{minipage}%
  }%
  \subfigure[Same-value Attack Loss]{
    \begin{minipage}[t]{0.3\linewidth}
      \centering
      \includegraphics[width=5.5cm]{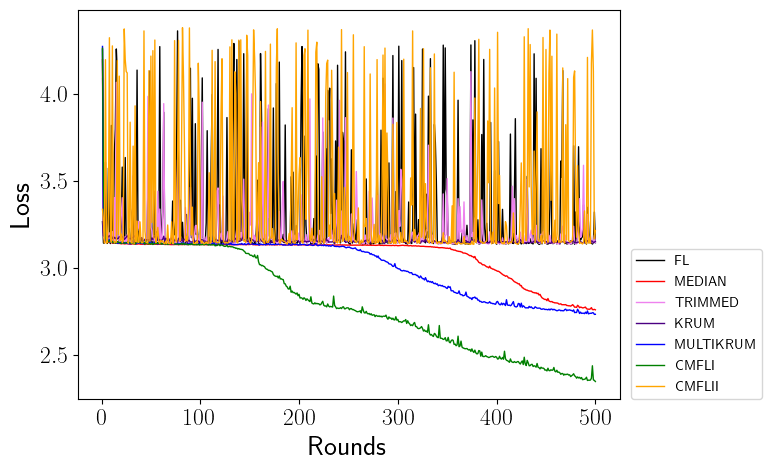}
    \end{minipage}%
  }%
  \subfigure[Back-gradient Attack Loss]{
    \begin{minipage}[t]{0.3\linewidth}
      \centering
      \includegraphics[width=5.5cm]{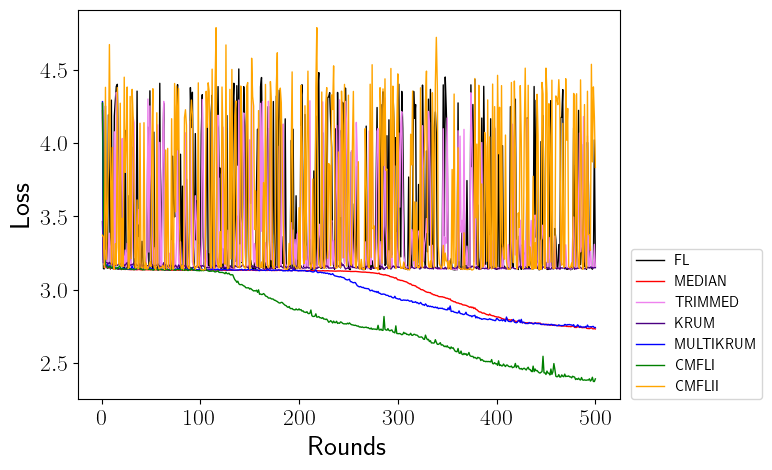}
    \end{minipage}%
  }%
  
  \subfigure[Gradient Scaling Attack Accuracy]{
    \begin{minipage}[t]{0.295\linewidth}
      \centering
      \includegraphics[width=5.5cm]{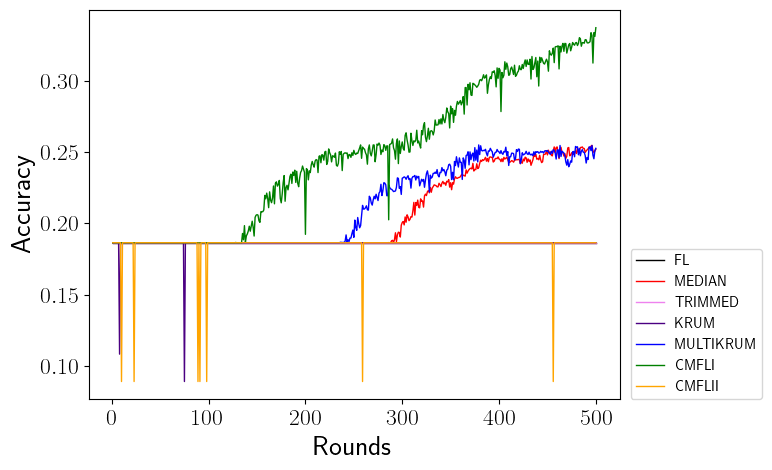}
    \end{minipage}
  }%
  \subfigure[Same-value Attack Accuracy]{
    \begin{minipage}[t]{0.295\linewidth}
      \centering
      \includegraphics[width=5.5cm]{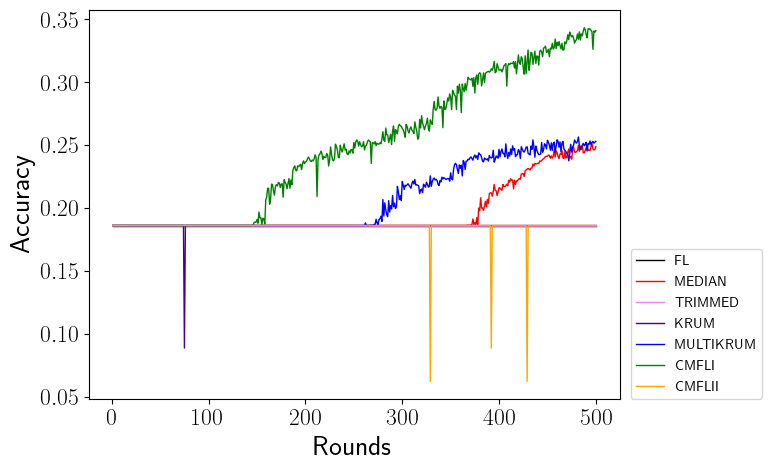}
    \end{minipage}
  }%
  \subfigure[Back-gradient Attack Accuracy]{
    \begin{minipage}[t]{0.295\linewidth}
      \centering
      \includegraphics[width=5.5cm]{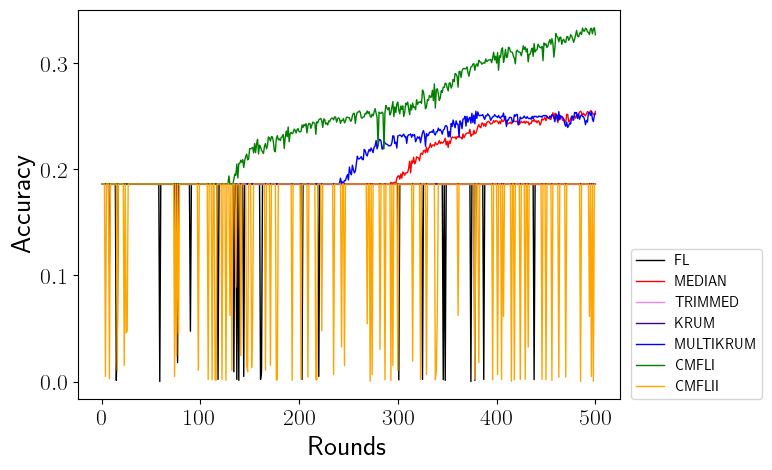}
    \end{minipage}%
  }%
  
  \centering
  \caption{Performance of CMFL compared with other Byzantine-tolerant algorithms under various attacks over Shakespeare dataset.}
  \label{fig:EXP1_Shakes}
\end{figure*}

We show the performance of CMFL compared with other Byzantine-tolerant algorithms under various attacks in Figure \ref{fig:EXP1_FEMNIST}-\ref{fig:EXP1_Shakes}. We analyzed the performance of each Byzantine-tolerant algorithm.

\begin{itemize}
\item \textbf{CMFL-selection strategy \uppercase\expandafter{\romannumeral1}} always reaches the fastest global convergence speed and highest accuracy among these five algorithms. Note that in the model accuracy curve, the CMFL curve fluctuates more obviously than Median and Multi-Krum, which is the normal curve fluctuation caused by the replacement of committee members. Nevertheless, its overall accuracy rate is still higher than the other algorithms.
\item \textbf{CMFL-selection strategy \uppercase\expandafter{\romannumeral2}} achieves as poor performance as typical FL. Obviously, the malicious clients can easily participate in the aggregation process.
\item \textbf{Median}\cite{pmlr-v80-yin18a} has maintained a relatively stable performance under a variety of attacks. It only selects the median of the local gradients as the global gradient, which is regarded as a relatively conservative approach. Although Median can effectively resist Byzantine attacks, it is difficult to achieve excellent model performance.
\item \textbf{Trimmed Mean}\cite{pmlr-v80-yin18a} has an unstable performance under different attacks. Trimmed Mean performs well under gradient scaling attack but still not as good as Median and Multi-Krum, and performed extremely badly under same-value attack and back-gradient attack. This Byzantine-tolerant algorithm cannot effectively resist the Byzantine attacks. 
\item \textbf{Krum}\cite{Blanchard2017Byzantine} performs poorly under all three attacks, and there were sharp fluctuations in the training process. Krum only selects the first local gradient after sorting as the global gradient each time. This aggregation method that does not consider the overall results in severe fluctuations of the global model.
\item \textbf{Multi-Krum}\cite{Blanchard2017Byzantine} has the second-best model performance overall. Multi-Krum improves the original Krum and eliminates the jitter generated during the training process. 
\end{itemize}

Note that the CMFL is the only framework without involving a central server, which naturally achieves robustness against the influence of a malicious server. Other Byzantine-tolerant algorithms cannot achieve the same robustness owing to their naturally centralized architecture.

\subsection{Hyper-parameter Analysis Experiment}\label{parameterAnalysisExperiment}

\subsubsection{Experiment Setting}

In this experiment, we consider the effect of hyper-parameter by varying the hyper-parameter $\alpha$, $\omega$ and $\epsilon$. Specifically, we consider the back-gradient attack and designed three sets of sub-experiments. In each set of sub-experiments, we fixed one of the hyper-parameters and changed the other two hyper-parameters to analyze their impacts on the model performance.

\begin{itemize}
	\item \textbf{Sub-experiment \uppercase\expandafter{\romannumeral1}.} Fixed $\alpha=40$ and vary $\omega, \epsilon$ to $\{10,20,30,40,50\}$.
	\item \textbf{Sub-experiment \uppercase\expandafter{\romannumeral2}.} Fixed $\omega=40$ and vary $\alpha, \epsilon$ to $\{10,20,30,40,50\}$.
	\item \textbf{Sub-experiment \uppercase\expandafter{\romannumeral3}.} Fixed $\epsilon=10$ and vary $\alpha, \omega$ to $\{10,20,30,40,50\}$.
\end{itemize}

\begin{figure*}[htbp]
	\centering
	
	\subfigure[$\alpha=40$]{
		\begin{minipage}[t]{0.3\linewidth}
			\centering
			\includegraphics[width=5.5cm]{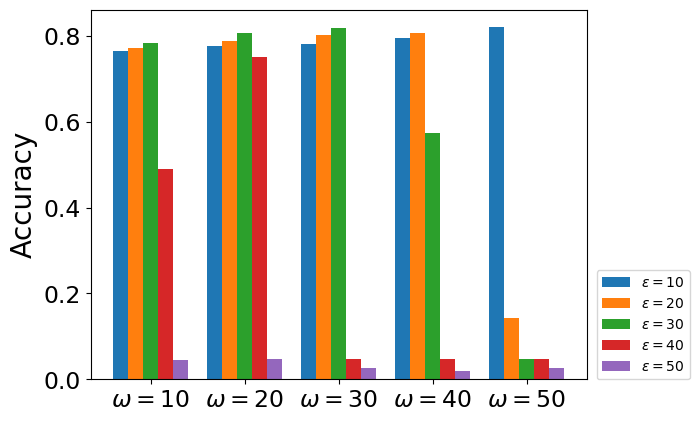}
		\end{minipage}%
	}%
	\subfigure[$\omega=40$]{
		\begin{minipage}[t]{0.3\linewidth}
			\centering
			\includegraphics[width=5.5cm]{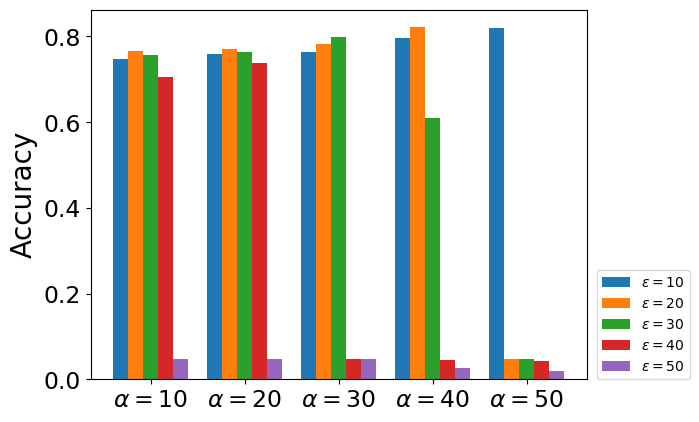}
		\end{minipage}%
	}%
	\subfigure[$\epsilon=10$]{
		\begin{minipage}[t]{0.3\linewidth}
			\centering
			\includegraphics[width=5.5cm]{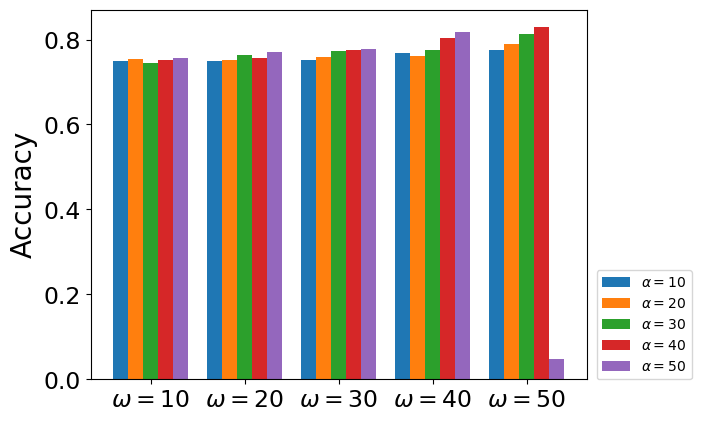}
		\end{minipage}%
	}%
	
	\centering
	\caption{Model performance of CMFL with varying the hyper-parameters.}
	\label{fig:EXP2}
\end{figure*}

	
	
	

\subsubsection{Result Analysis}
 
We show the results in Figure \ref{fig:EXP2} and the analysis as follows:

\begin{itemize}
\item \textbf{Appropriately increasing the number of committee members can enhance the performance of the global model.} The results show that within a suitable parameter value range (e.g., $\alpha=10, \epsilon \leq 30, \omega \leq 30$), more committee members lead to better global model performance. This is mainly since an appropriate increase in the number of committee members can enhance the robustness of the committee to a certain extent, in the meanwhile avoid the existence of "dictators" and prevent the training process from being controlled by a small number of clients, which makes it difficult for some other clients to participate in the aggregation process. 
\item \textbf{Appropriately increasing the proportion of the aggregation clients can enhance the performance of the global model.} Within a suitable parameter value range, a higher proportion of the aggregation clients leads to better global model performance. Because in the normal operation of the committee, sorting the local gradients uploaded by the training client according to their scores makes the honest local gradients come first and the malicious gradients second. In this ideal situation, we control the proportion of aggregated clients to be less than a threshold $\chi$ to prevent malicious gradients from participating in the aggregation process. Under the limit of $\alpha \in (0,\chi]$, the increase of $\alpha$ means that more honest local gradients are selected to construct the global gradient in each round, which can achieve better global model performance.
\item \textbf{Excessive $\alpha$, $\omega$, and $\epsilon$ will lead to a cliff-like decline in the performance of the global model.} When we increase the proportion of committee members, it means that both malicious clients and honest clients have a greater probability of being elected as committee members. This increases the probability of malicious clients mixing into committee members and damage the committee’s scoring system, making it difficult to assign the correct score to the training clients. In the worst situation, once the proportion of the malicious clients among committee members is relatively large, the scores of the malicious clients will be higher than those of the honest clients, resulting in the malicious local gradient uploaded by the malicious clients being used to construct the global gradient. Hence, the performance of the global model will be devastatingly damaged. When we increase the proportion of the aggregation clients, it also means increasing the probability of malicious local gradients participating in the aggregation procedure. When the proportion of aggregation clients exceeds the threshold that the system can tolerate, the malicious local gradients are used to construct the global gradient, which causes irreparable damage to the performance of the global model. By the same token, if there are too many malicious clients, the malicious local gradients uploaded by them are more likely to participate in the aggregation process. As long as a malicious local gradient is aggregated, the performance of the global model will be greatly reduced.
\end{itemize}

\subsection{Efficiency Experiment}
  
In this section, we evaluate the efficiency of CMFL over FEMNIST and Sentiment140 dataset. Besides, we compare it with other decentralized FL frameworks. 

\subsubsection{Experiment Settting}

\textbf{Implementation Detail}. In this experiment, we simulate the data transferring by calculating the transmission time using $T_{transmission} = s/r$, where $s$ represents the data and $r$ represents the transmission rate. We let the program process wait for $T_{transmission}$ second when it needs to transmit data for simulating the data transferring in a real scene. The maximum number of communication rounds is 600. 

\textbf{Hyper-parameter Setting}. In order to exclude the influence of irrelevant variables, we conduct this evaluation without considering the Byzantine attacks. The number of committee clients and training clients is set as 43 and 65. And the aggregation rate $\alpha$ is set as $40\%$. We set different maximum transmission rates for one client: $1Mps$, $10Mps$, and $100Mps$, and under each transmission rate, we evaluate the performance of CMFL using wall-clock time, which includes computation time and communication time. 

\textbf{Comparative Method}. We compare the efficiency of CMFL with three algorithms: typical FL, BrainTorrent\cite{DBLP:journals/corr/abs-1905-06731}, and GossipFL\cite{DBLP:journals/corr/abs-1908-07782}.

\subsubsection{Result Analysis}
\begin{figure*}[htbp]
  \centering
  
  \subfigure[FEMNIST-1Mps]{
    \begin{minipage}[t]{0.3\linewidth}
      \centering
      \includegraphics[width=5.5cm]{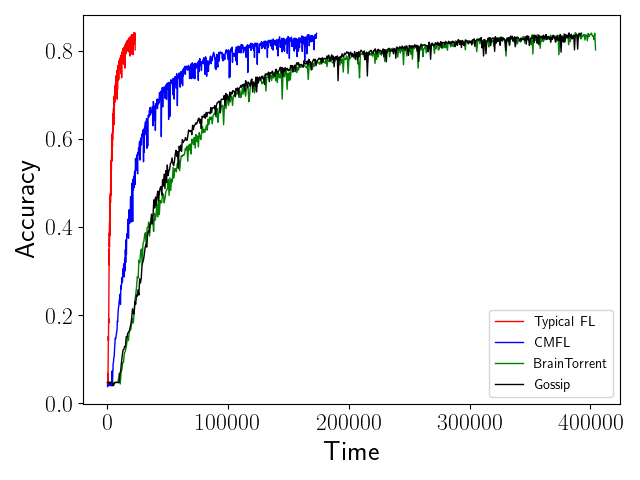}
    \end{minipage}%
  }%
  \subfigure[FEMNIST-10Mps]{
    \begin{minipage}[t]{0.3\linewidth}
      \centering
      \includegraphics[width=5.5cm]{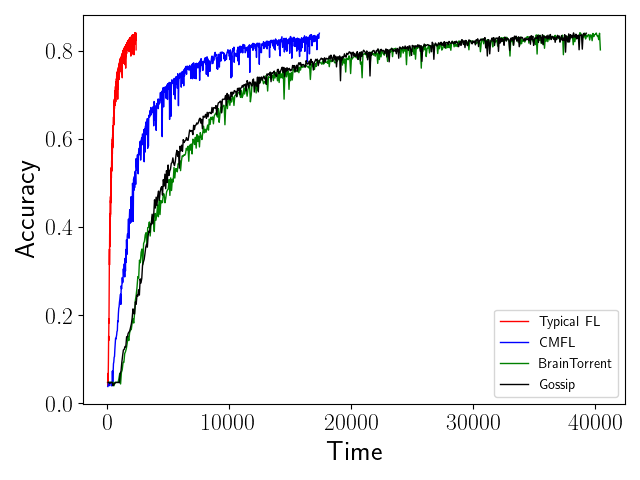}
    \end{minipage}%
  }%
  \subfigure[FEMNIST-100Mps]{
    \begin{minipage}[t]{0.3\linewidth}
      \centering
      \includegraphics[width=5.5cm]{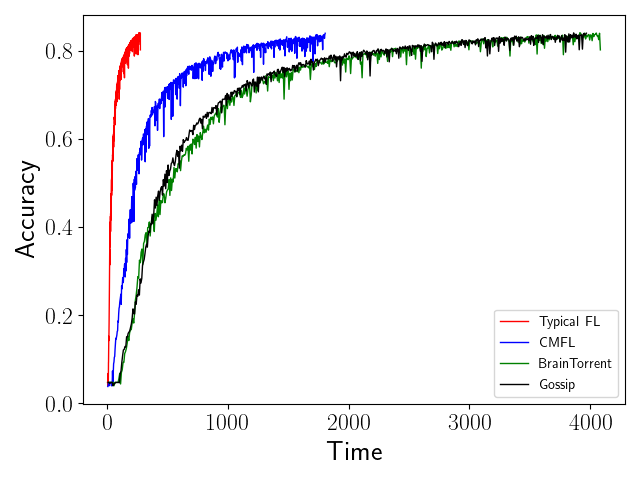}
    \end{minipage}%
  }%
  
  \subfigure[Sentiment140-1Mps]{
    \begin{minipage}[t]{0.295\linewidth}
      \centering
      \includegraphics[width=5.5cm]{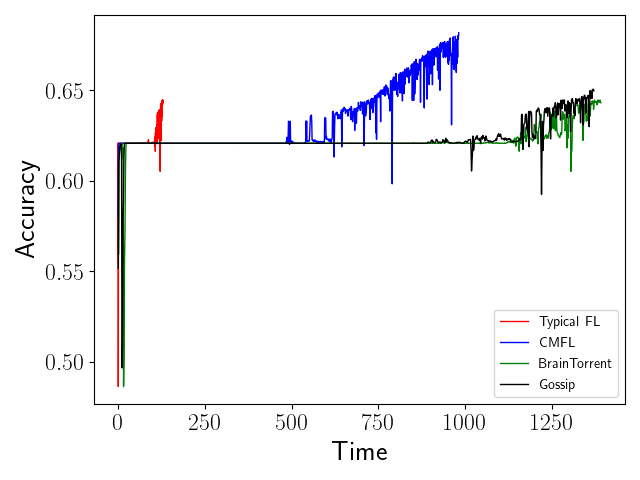}
    \end{minipage}
  }%
  \subfigure[Sentiment140-10Mps]{
    \begin{minipage}[t]{0.295\linewidth}
      \centering
      \includegraphics[width=5.5cm]{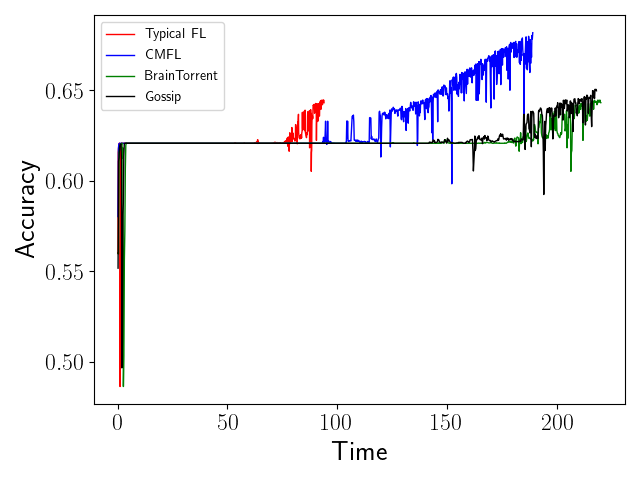}
    \end{minipage}
  }%
  \subfigure[Sentiment140-100Mps]{
    \begin{minipage}[t]{0.295\linewidth}
      \centering
      \includegraphics[width=5.5cm]{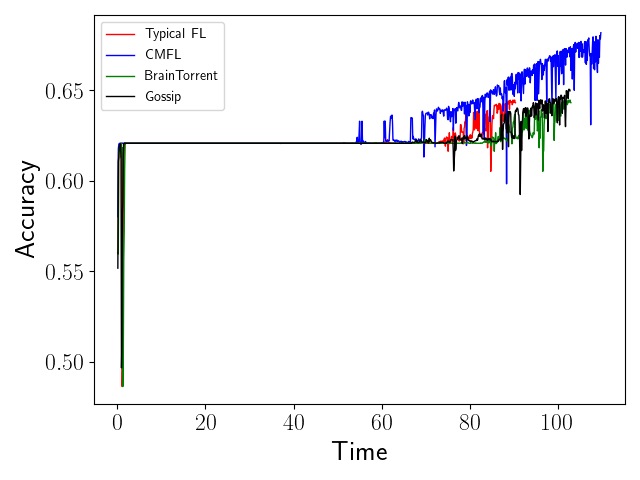}
    \end{minipage}%
  }%
  
  \centering
  \caption{Performance of typical FL and three decentralized FL frameworks under different transmission rates over FEMNIST dataset and Sentiment140 dataset.}
  \label{fig:communication}
\end{figure*}

\begin{figure*}[htbp]
  \centering
  
  
  \subfigure[Sentiment140-1Mps]{
    \begin{minipage}[t]{0.295\linewidth}
      \centering
      \includegraphics[width=5.5cm]{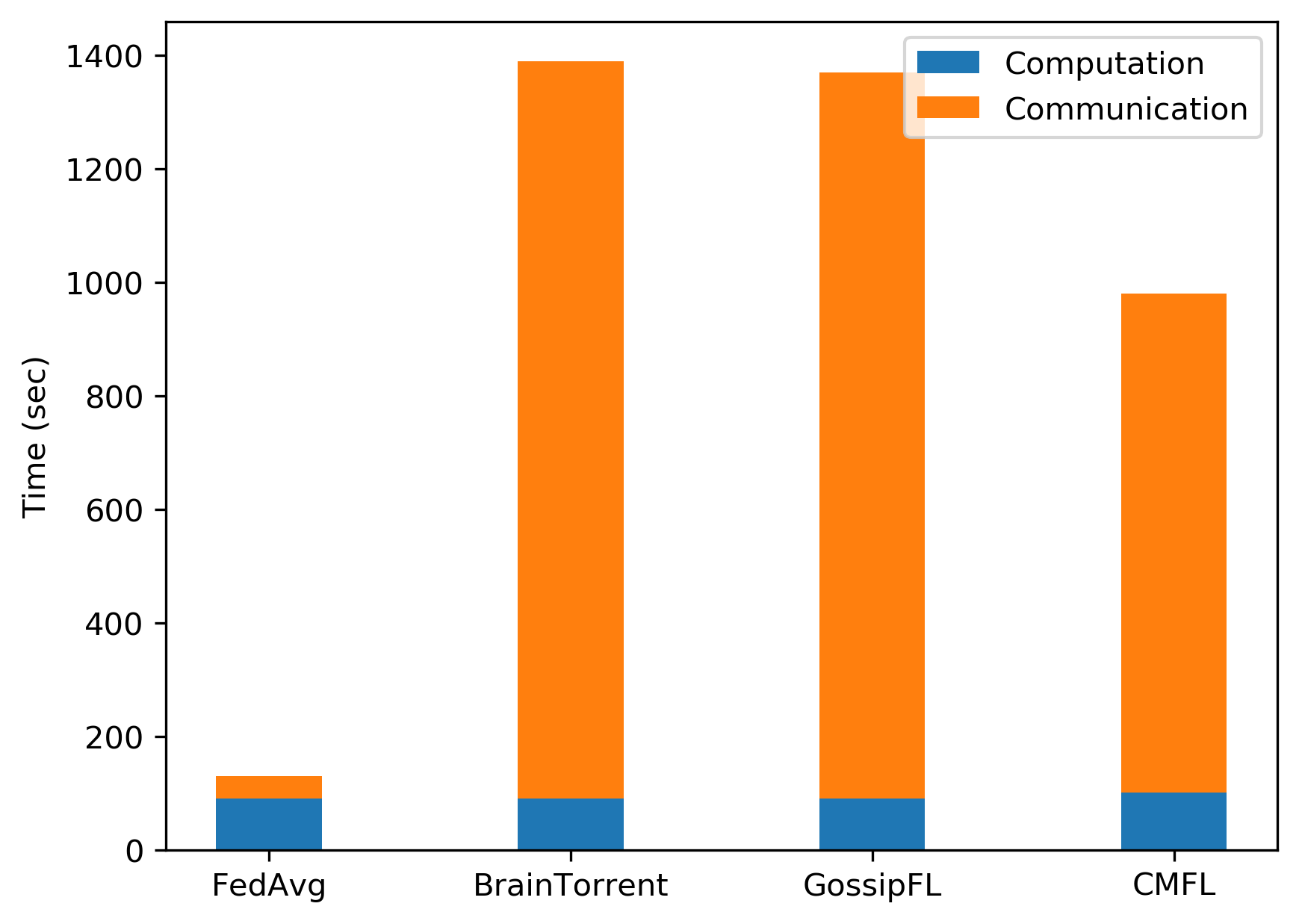}
    \end{minipage}
  }%
  \subfigure[Sentiment140-10Mps]{
    \begin{minipage}[t]{0.295\linewidth}
      \centering
      \includegraphics[width=5.5cm]{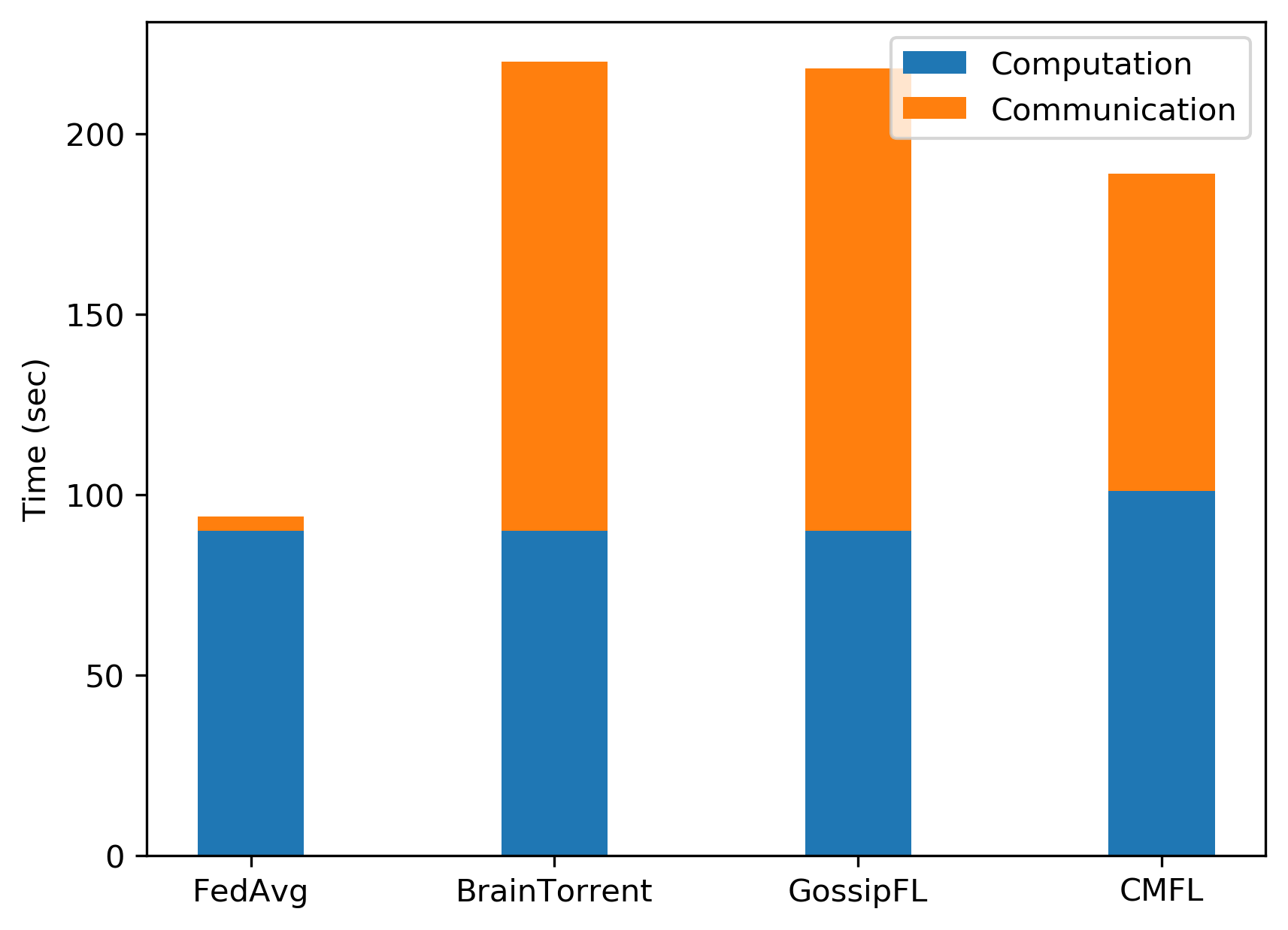}
    \end{minipage}
  }%
  \subfigure[Sentiment140-100Mps]{
    \begin{minipage}[t]{0.295\linewidth}
      \centering
      \includegraphics[width=5.5cm]{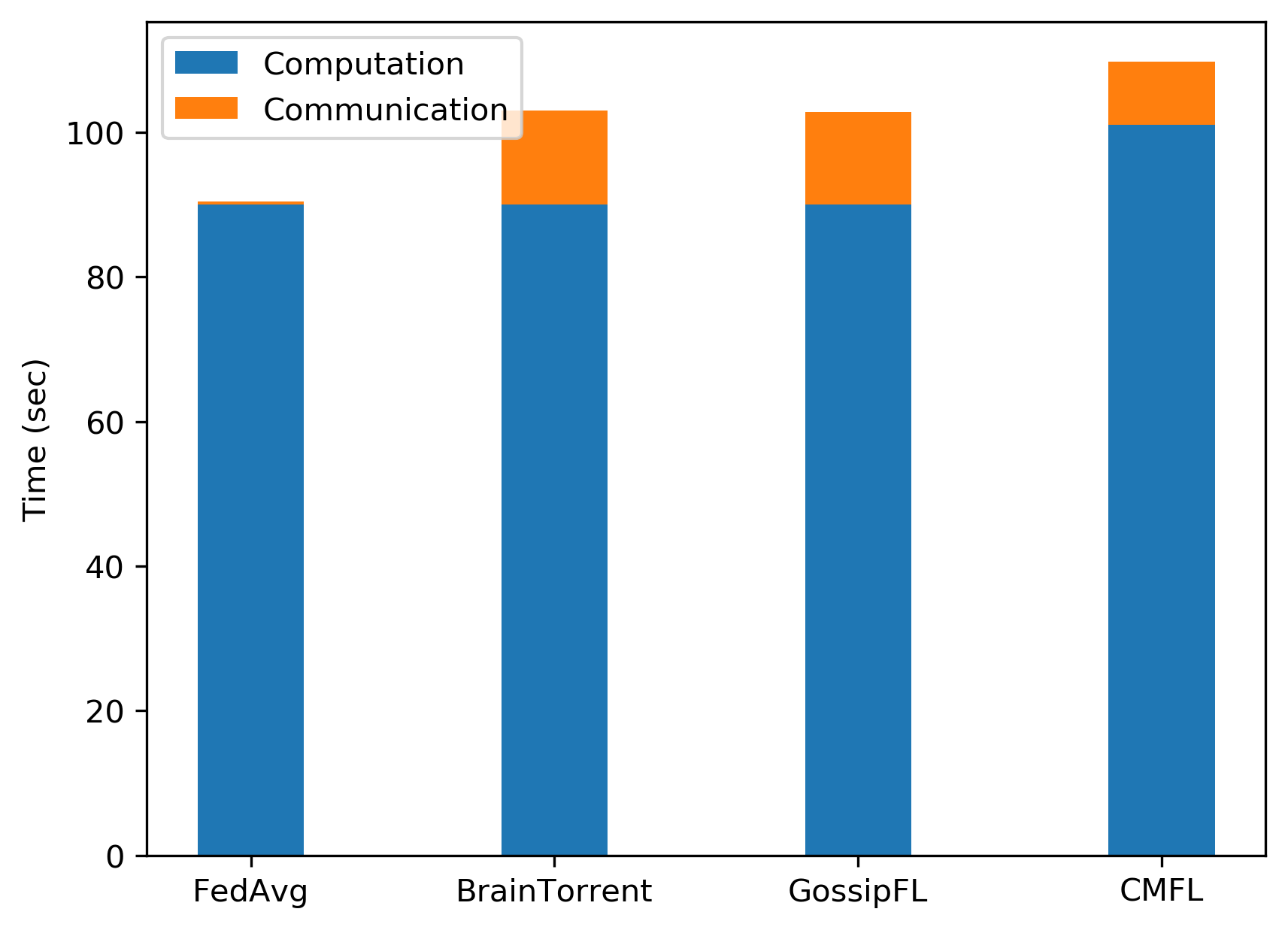}
    \end{minipage}%
  }%
  
  \centering
  \caption{The total communication time and computation time of typical FL and three decentralized FL frameworks over the Sentiment140 dataset.}
  \label{fig:communication_computation}
\end{figure*}

{
Figure \ref{fig:communication} shows the result. Noted that the size of AlexNet ($\approx 200$M) is much bigger than that of LSTM ($\approx 16$k), so the communication time over FEMNIST dataset is much longer than that over the Sentiment140 dataset. The overall performance of CMFL is better than the other two decentralized FL frameworks but worse than typical FL. Typical FL has an overall better performance than CMFL because the clients do not have to communicate with each other. It is a centralized framework so the clients just send their local gradients to a server for aggregation. Never can the other three decentralized frameworks do that because the client has to broadcast their local gradients to other clients for reaching consensus, in which case the transmission of the local gradients occurs communication overhead. With the increasing transmission rate, the advantage of typical FL becomes smaller. The performance of the typical FL is taken over by CMFL when the transmission is up to $100Mps$ over the Sentiment140 dataset. CMFL always has a better performance than the other two decentralized frameworks, because it utilized the committee system to reduce the communication overhead, which is proved in Figure \ref{fig:communication_computation}. The clients do not need to broadcast their local gradients to other clients but send their local gradients to committee clients. The committee clients can reach a consensus with lower communication overhead by performing CCP, while the other two decentralized frameworks must cost higher communication overhead. Considering the malicious server in a real scenario, CMFL can achieve both robustness and efficiency. 
}

\subsection{Committee Member Analysis Experiment}\label{CommitteeAnalysisEvaluation}

\subsubsection{Experiment Setting}

In this experiment we set $\alpha = 40, \omega = 30$ and vary $\epsilon$ to $\{10,20,30,40,50\}$. By recording the number of malicious training clients, malicious committee clients and malicious aggregation clients during the training process, we analyze the influence of committee members on the performance of the global model. 

\begin{figure}
	\centering
	\includegraphics[width=7cm]{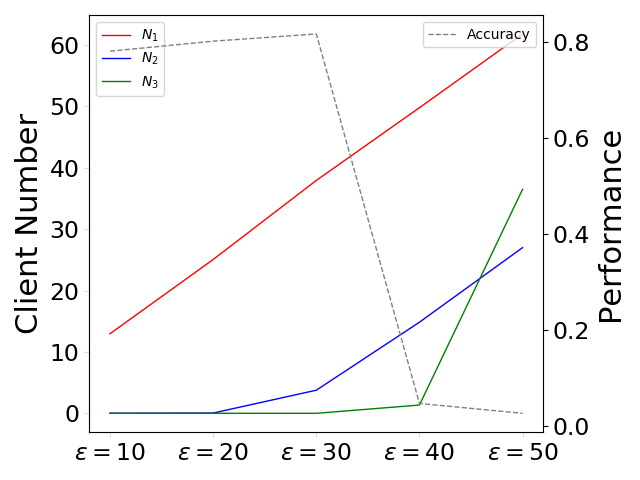}
	\caption{Number of malicious clients in the training process, where $N_1$ denotes the number of malicious training clients, $N_2$ denotes the malicious committee clients and $N_3$ denotes the malicious aggregation clients.}
	\label{fig:malicious_committe}
\end{figure}

\subsubsection{Result Analysis}

We show the result in Figure \ref{fig:malicious_committe}. Since in each round the training clients are randomly selected from the non-committee clients, when $\epsilon$ increases, the number of malicious training clients also increases. And with the increase of $\epsilon$, some malicious clients will inevitably be mixed into the committee members. Nevertheless, a small number of miscellaneous malicious clients cannot destroy the entire committee's scoring system and influence the committee's judgment, which instead improve the performance of the global model. This is mainly because the scoring system does not lose the ability to distinguish between the malicious clients and the honest clients since those malicious clients still receive a score much lower than that of honest clients. But for the honest clients, they get almost the same score due to the extreme scores assigned by the malicious committee clients. In this case, our method is equivalent to the typical FL method without considering the Byzantine attack. The typical FL achieves better performance than CMFL in such a setting, which has shown in Experiment \ref{nomaltrainingexperiment}. However, if there are too many mixed malicious clients, the committee's scoring system will be destroyed and the committee members will lose the ability to eliminate malicious local gradients, resulting in a sharp drop in the performance of the global model.  

\section{Conclusion}
\label{conclusion}

In this paper, we propose a serverless FL framework under committee mechanism, which can ensure robustness when considering Byzantine attacks. Besides, we present the convergence guarantees for our proposed framework. Motivated by the insight from the theoretical analysis we design the election and selection strategies, which empower the model the robustness against both the Byzantine attack and malicious server problem. The experiment results demonstrate the outperformance of the model over the typical federated learning and Byzantine-tolerant models, which further verify the effectiveness and robustness of the proposed framework. 

Currently, the proposed framework mainly ensures the robustness of the aggregation procedure by detecting the abnormal local gradients. While in the face of targeted attacks, such as a backdoor attack, how to design suitable election and selection strategies with theoretical guarantee remains an open and worth-exploring topic in the future.


%


\ifCLASSOPTIONcompsoc
  \section*{Acknowledgments}
\else
  \section*{Acknowledgment}
\fi

The research is supported by the National Key R\&D Program of China (2020YFB1006001), the National Natural Science Foundation of China (62176269, 62006252), the Key-Area Research and Development Program of Guangdong Province (2020B010165003), and the Innovative Research Foundation of Ship General Performance (25622112).

\ifCLASSOPTIONcaptionsoff
  \newpage
\fi



%



\bibliographystyle{IEEEtran}
\bibliography{ref}

%

\begin{IEEEbiography}
 [{\includegraphics[width=1in,height=1.25in,clip,keepaspectratio]{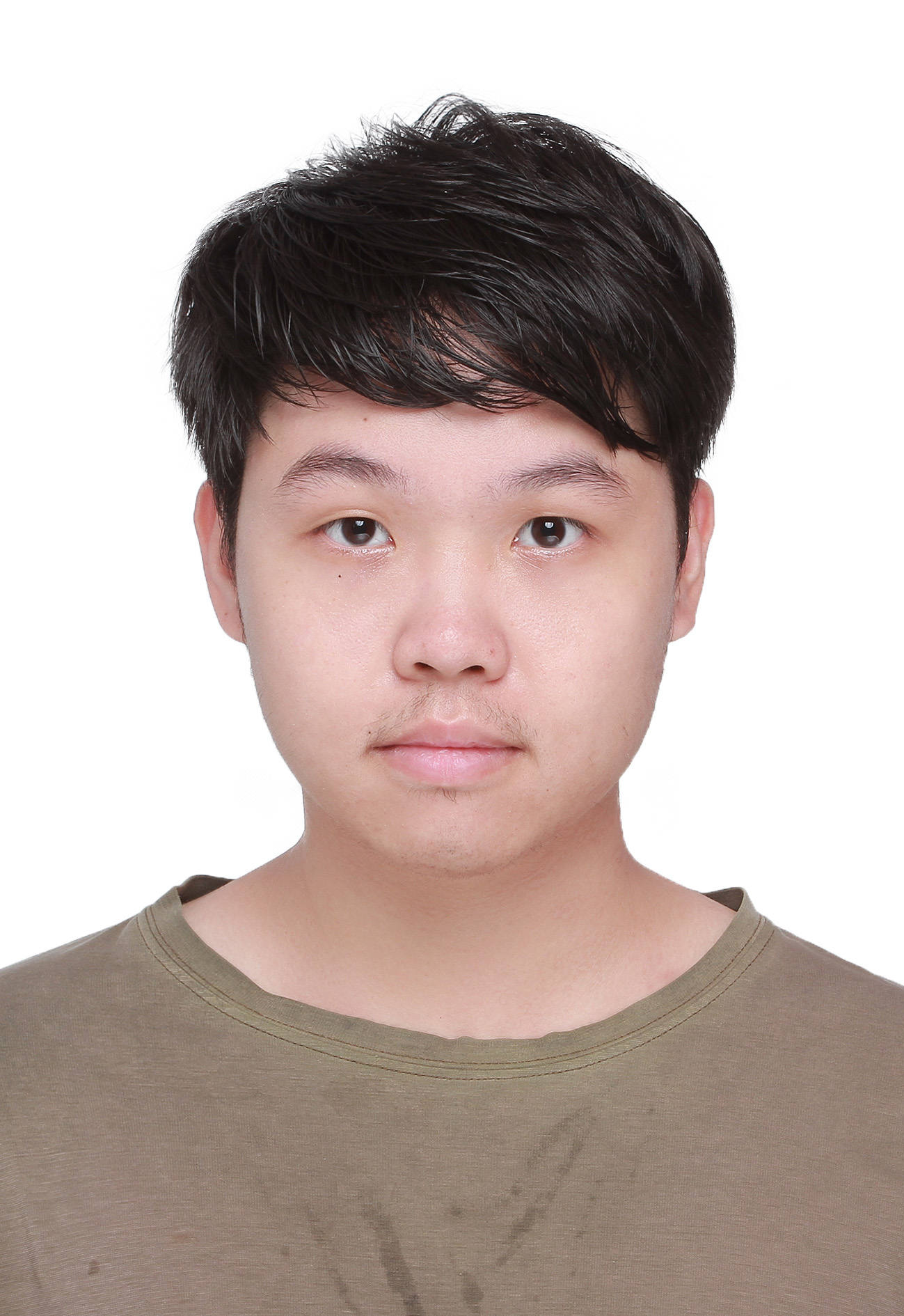}}]{Chunjiang Che} is currently working toward the B.E. degree in the School of Computer Science, Sun Yat-sen University, Guangzhou, China. His recent research interests include blockchain, federated learning and optimization.
\end{IEEEbiography}
\begin{IEEEbiography}
 [{\includegraphics[width=1in,height=1.25in,clip,keepaspectratio]{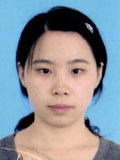}}]{Xiaoli Li} is currently working toward the PhD degree in the School of Computer Science and Engineering, Sun Yat-sen University, Guangzhou, China. She received the Master's degree in computer architecture from University of Electronic Science and Technology of China, Chengdou, China, in 2011. Her research interests include services computing, software engineering, cloud computing, machine learning and federated learning.
\end{IEEEbiography}

\begin{IEEEbiography}
 [{\includegraphics[width=1in,height=1.25in,clip,keepaspectratio]{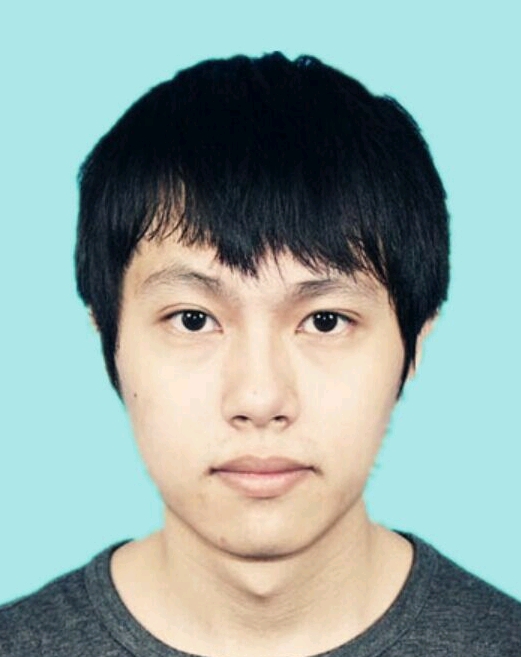}}]{Chuan Chen} received the B.S. degree from Sun Yat-sen University, Guangzhou, China, in 2012, and the Ph.D. degree from Hong Kong Baptist University, Hong Kong, in 2016. He is currently an Associate Professor with the School of Computer Science and Engineering, Sun Yat-Sen University. He published over 50 international journal and conference papers. His current research interests include machine learning, numerical linear algebra, and numerical optimization.
\end{IEEEbiography}
\begin{IEEEbiography}
 [{\includegraphics[width=1in,height=1.25in,clip,keepaspectratio]{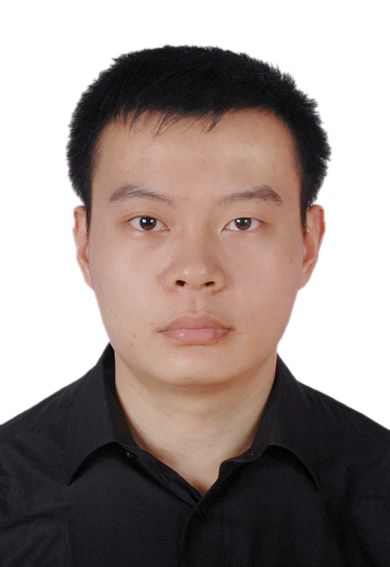}}]{Xiaoyu He}
received the B.Eng. degree in computer science and technology from Beijing Electronic Science and Technology Institute, Beijing, China, in 2010, the M.Sc. degree in public administration from South China University of Technology, Guangzhou, China, in 2016, and the Ph.D. degree in computer science from Sun Yat-sen University, Guangzhou, in 2019. He is currently a postdoctoral fellow at School of Data and Computer Science, Sun Yat-sen University. His research interests include evolutionary computation and machine learning.
\end{IEEEbiography}
\begin{IEEEbiography}
  [{\includegraphics[width=1in,height=1.25in,clip,keepaspectratio]{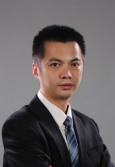}}]{Zibin Zheng} received his PhD degree from the Chinese University of Hong Kong in 2011. He is currently a professor in the School of Computer Science and Engineering at Sun Yat-sen University, China. He has published over 150 international journal and conference papers, including three ESI highly cited papers. According to Google Scholar, his papers have more than 13,590 citations, with an H-index of 54. His research interests include blockchain, smart contract, services computing, software reliability.
\end{IEEEbiography}




\onecolumn
\appendices
\renewcommand\thesection{\Alph{section}}

\section{Proof of the Theorem 1}
\label{proof}

\subsection{Definition and Lemma}

In our analysis, the $(t)$ is defined as the index of the total local SGD iterations, where $(i) = t\tau + i$. We iterpret $\w_k^{(i)}$ as the local model of $k$-th client at iteration $(t)$. In such a setting, the aggregation client set and the commitee client set are expressed as $S_a^{(t)}$ and $S_c^{(t)}$. Note that the aggregation clients and committee clients perform $\tau$ iterations of local SGD, the aggregation client set $S_a^{(t)}$ and the committee client set $S_c^{(t)}$ remain the same for every $\tau$ iterations. That is, $S_a^{(t)} = S_a^{(t+1)} = ...=S_a^{(t+\tau-1)}$ and $S_c^{(t)} = S_c^{(t+1)} = ...=S_c^{(t+\tau-1)}$, where $t \% \tau = 0$. The local gradient of $k$-th client is expressed as $g_k(\w_k^{(t)},\B_{k}^{(t)})$, which is used for the local SGD training:
\begin{equation}
\begin{split}
&\w_k^{(t+1)} = \w_k^{(t)} - \eta_tg_k(\w_k^{(t)},\B_{k}^{(t)}),
\end{split}
\end{equation}where $\w_k^{(t)}$ denotes the local model of the $k$-th client at iteration $t$. Actually we only update the global model $\overline{\w}^{(t)}$ after every $\tau$ rounds, but for the need of convergence proof and analysis we assume the $\overline{\w}^{(t)}$ is updated at each iteration as follows:
\begin{equation}\label{globalUpdate2}
\begin{split}
&\overline{\w}^{(t+1)} = \overline{\w}^{(t)} - \eta_t\overline{g}^{(t)} = \overline{\w}^{(t)} - \eta_t(\sum_{k\in S^{(t)}_a}p_{k,S_a^{(t)}}g_k(\w_k^{(t)},\B_k^{(t)})),
\end{split}
\end{equation}where $\eta_t$ is the learning rate at iteration $(t)$. At iteration $(t)$ which satisfies $t\%\tau=0$, the $k$-th client downloads the $\overline{\w}^{(t)}$ to the local as the new local model $\w_k^{(t)}$, as a result, the whole updating method of the local model can be written as follows:

\begin{equation}
\label{localUpdate}
\w_k^{(t+1)} = \left\{
\begin{aligned}
& \w_k^{(t)} - \eta_t g_k(\w_k^{(t)},\B_k^{(t)}), (t+1) \% \tau \neq 0 \\
&\overline{\w}^{(t+1)}, (t+1) \% \tau = 0.
\end{aligned}
\right.
\end{equation}Then we introduce our proposed lemmas.
\begin{lemma}\label{l1}
\textbf{(Global-Local Gradient Product). } \textit{According to the definitions of $\nabla F_k(\w_k^{(t)})$, we have that}
\begin{equation}
\begin{split}
&-\langle \overline{\w}^{(t)}-\w^*,\sum_{k\in \s}p_{k,\s}\nabla F_k(\w_k^{(t)})\rangle \\
\leq& \frac{1}{2\eta_t}\sum_{k \in \s}p_{k,\s}||\overline{\w}^{(t)} - \w_k^{(t)}||^2 + \frac{\eta_t}{2}\sum_{k\in S_a^{(t)}}p_{k,\s}||\nabla F_k(\w_k^{(t)})||^2 - \sum_{k \in \s}p_{k,\s}(F_k(\w_k^{(t)})-F_k(\w^*))\\
&- \frac{\mu}{2}\sum_{k \in \s}p_{k,\s}||\w_k^{(t)}-\w^*||^2. 
\end{split}
\end{equation}
\end{lemma}
\begin{lemma}\label{l2}
\textbf{(Local Parameter-Optimal Gap).} \textit{Let $v_t = 2\eta_t-4L\eta_t^2 = 2\eta_t(1-2L\eta_t)$, we have that}
\begin{equation}
\begin{split}
&-\sum_{k \in \s}p_{k,\s}(F_k(\w_k^{(t)}) - F_k^*)\\
\leq& -(1-\eta_tL)\sum_{k \in \s}p_{k,\s}(F_k(\overline{\w}^{(t)}) - F^*)+ \frac{1}{v_t}\sum_{k\in \s}p_{k,\s}||\w_k^{(t)} - \overline{\w}^{(t)}||^2.
\end{split}
\end{equation}
\end{lemma}

\begin{lemma}\label{l4}
\textbf{(Client Heterogeneous Bound).} \textit{According to the Assumption 4, we have that}
\begin{equation}
\begin{split}
&\mathbb{E}[\sum_{k \in \s}p_{k,\s}||\overline{\w}^{(t)}-\w_k^{(t)}||^2] \leq 16\eta_t^2\tau^2G^2.
\end{split}
\end{equation}
\end{lemma}

\subsection{Proof of Lemma}

In this section we will prove Lemma \ref{l1} to \ref{l4}. 
\subsubsection{Proof of Lemma \ref{l1}}
\begin{proof}
We introduce $\w_k^{(t)}$ into the formula:
\begin{equation}
\begin{split}
&-\langle \overline{\w}^{(t)}-\w^*,\sum_{k\in \s}p_{k,\s}\nabla F_k(\w_k^{(t)})\rangle \\
=& -\sum_{k\in \s}\langle\overline{\w}^{(t)}-\w^*,p_{k,\s}\nabla F_k(\w_k^{(t)})\rangle \\
=& -\sum_{k \in \s}p_{k,\s}\langle\overline{\w}^{(t)} - \w_k^{(t)} + \w_k^{(t)} - \w^*,\nabla F_k(\w_k^{(t)})\rangle \\
=& -\sum_{k \in \s}p_{k,\s}\langle\overline{\w}^{(t)} - \w_k^{(t)},\nabla F_k(\w_k^{(t)})\rangle -\sum_{k \in \s}p_{k,\s}\langle\w_k^{(t)}-\w^*,\nabla F_k(\w_k^{(t)})\rangle.
\end{split}
\end{equation}
According to the Cauchy-Schwarz inequality and AM-GM inequality, we have
\begin{equation}
\begin{split}
&-\langle\overline{\w}^{(t)}-\w^*,\sum_{k\in \s}p_{k,\s}\nabla F_k(\w_k^{(t)})\rangle \\
\leq& \frac{1}{2}\sum_{k \in \s}p_{k,\s}(\frac{1}{\eta_t}||\overline{\w}^{(t)}-\w_k^{(t)}||^2 + \eta_t||\nabla F_k(\w_k^{(t)})||^2)-\sum_{k \in \s}p_{k,\s}\langle\w_k^{(t)}-\w^*,\nabla F_k(\w_k^{(t)})\rangle\\
\leq& \frac{1}{2}\sum_{k \in \s}p_{k,\s}(\frac{1}{\eta_t}||\overline{\w}^{(t)}-\w_k^{(t)}||^2 + \eta_t||\nabla F_k(\w_k^{(t)})||^2)-\sum_{k \in \s}p_{k,\s}(\w_k^{(t)}-\w^*)^T\nabla F_k(\w_k^{(t)}).
\end{split}
\end{equation}
Due to the Assumption 2, we expand the above formula and complete the proof:
\begin{equation}
\begin{split}
&-\langle\overline{\w}^{(t)}-\w^*,\sum_{k\in \s}p_{k,\s}\nabla F_k(\w_k^{(t)})\rangle \\
\leq&  \frac{1}{2}\sum_{k \in \s}p_{k,\s}(\frac{1}{\eta_t}||\overline{\w}^{(t)}-\w_k^{(t)}||^2 + \eta_t||\nabla F_k(\w_k^{(t)})||^2)- \sum_{k \in \s}p_{k,\s}(F_k(\w_k^{(t)}-F_k(\w^*))-\frac{\mu}{2}||\w_k^{(t)} - \w^*||^2) \\
=& \frac{1}{2\eta_t}\sum_{k \in \s}p_{k,\s}||\overline{\w}^{(t)} - \w_k^{(t)}||^2 + \frac{\eta_t}{2}\sum_{k\in S_a^{(t)}}p_{k,\s}||\nabla F_k(\w_k^{(t)})||^2 - \sum_{k \in \s}p_{k,\s}(F_k(\w_k^{(t)})-F_k(\w^*))\\
&- \frac{\mu}{2}\sum_{k \in \s}p_{k,\s}||\w_k^{(t)}-\w^*||^2. 
\end{split}
\end{equation}
\end{proof}
\subsubsection{Proof of Lemma \ref{l2}}
\begin{proof}
The original formula can be written as:
\begin{equation}
\begin{split}
&-\sum_{k \in \s}p_{k,\s}(F_k(\w_k^{(t)}) - F_k^*) \\
=& -\sum_{k \in \s}p_{k,\s}(F_k(\w_k^{(t)}) - F_k(\overline{\w}^{(t)}) + F_k(\overline{\w}^{(t)}) - F_k^*) \\
\leq& -\sum_{k \in \s}p_{k,\s}(F_k(\w_k^{(t)}) - F_k(\overline{\w}^{(t)})) - \sum_{k \in \s}p_{k,\s}(F_k(\overline{\w}^{(t)}) - F_k^*).
\end{split}
\end{equation}
Due to the assumption 2 we have
\begin{equation}
\begin{split}
&-\sum_{k \in \s}p_{k,\s}(F_k(\w_k^{(t)}) - F_k^*) \\
\leq& \sum_{k \in \s}p_{k,\s}[\frac{\eta_t}{2}||\nabla F_k(\overline{\w}^{(t)})||^2 + \frac{1}{2\eta_t}||\w_k^{(t)} - \overline{\w}^{(t)}||^2 - \frac{\mu}{2}||\w_k^{(t)}-\overline{\w}^{(t)}||^2 - (F_k(\overline{\w}^{(t)})-F_k^*)]. 
\end{split}
\end{equation}
We continue to expand the above formula as follows:
\begin{equation}
\begin{split}
&-\sum_{k \in \s}p_{k,\s}(F_k(\w_k^{(t)}) - F_k^*) \\
=& \frac{\eta_t}{2}\sum_{k \in \s}p_{k,\s}||\nabla F_k(\overline{\w}^{(t)})||^2 - \sum_{k \in \s}p_{k,\s}(F_k(\overline{\w}^{(t)}) - F_k^*) + (\frac{1}{2\eta_t} - \frac{\mu}{2})\sum_{k \in \s}p_{k,\s}||\w_k^{(t)} - \overline{\w}^{(t)}||^2. 
\end{split}
\end{equation}
Due to the Assumption 1, we have
\begin{equation}
\begin{split}
&-\sum_{k \in \s}p_{k,\s}(F_k(\w_k^{(t)}) - F_k^*) \\
\leq& \eta_tL\sum_{k \in \s}p_{k,\s}(F_k(\overline{\w}^{(t)}) - F_k^*) - \sum_{k \in \s}p_{k,\s}(F_k(\overline{\w}^{(t)}) - F_k^*) + \frac{1-\eta_t\mu}{2\eta_t}\sum_{k \in \s}p_{k,\s}||\w_k^{(t)} - \overline{\w}^{(t)}||^2 \\
=& -(1-\eta_tL)\sum_{k \in \s}p_{k,\s}(F_k(\overline{\w}^{(t)}) - F^*) + \frac{1-\eta_t\mu}{2\eta_t}\sum_{k \in \s}p_{k,\s}||\w_k^{(t)} - \overline{\w}^{(t)}||^2.
\end{split}
\end{equation}
As $\frac{1-\eta_tL}{2\eta_t} \leq \frac{1}{v_t}$, we continue to expand the formula and complete the proof:
\begin{equation}
\begin{split}
&-\sum_{k \in \s}(F_k(\w_k^{(t)}) - F_k^*) \\
\leq& -(1-\eta_tL)\sum_{k \in \s}p_{k,\s}(F_k(\overline{\w}^{(t)}) - F^*) + \frac{1}{v_t}\sum_{k\in \s}p_{k,\s}||\w_k^{(t)} - \overline{\w}^{(t)}||^2.
\end{split}
\end{equation}
\end{proof}

\subsubsection{Proof of Lemma \ref{l4}}
\begin{proof}
According to the update rule, for $k$ and $k'$ which are in the same set $\s$, the term $||\w_{k'}^{(t)}-\w_k^{(t)}||^2$ will be zero when $k = k'$. As a result we have
\begin{equation}
\label{eq32}
\begin{split}
&\sum_{k\in \s}p_{k,\s}||\overline{\w}^{(t)}-\w_k^{(t)}||^2 \\
=& \sum_{k\in \s}p_{k,\s}||\sum_{k' \in \s}p_{k',\s}\w_{k'}^{(t)}-\w_k^{(t)}||^2 \\
=& \sum_{k\in \s}p_{k,\s}||\sum_{k' \in \s}(p_{k',\s}\w_{k'}^{(t)}-p_{k',\s}\w_k^{(t)})||^2 \\
=& \sum_{k\neq k' \atop k,k'\in \s}p_{k,\s}p_{k',\s}||\w_{k'}^{(t)}-\w_k^{(t)}||^2.
\end{split}
\end{equation}Since the selected local models are udpated with the global model at every $\tau$, for any $t$ there is a $t_0$ satisfies that $0 \leq t-t_0 \leq \tau$ and $\w_{k'}^{(t_0)} = \w_k^{(t_0)} = \overline{\w}^{(t)}$. Therefore for any $(t)$ the term $||\w_{k'}^{(t)}-\w_k^{(t)}||^2$ can be bound by $\tau$ epochs. With non-increasing $\eta_t$ over $(t)$ and $\eta_{t_0} \leq 2\eta_t$, Eq. \ref{eq32} can be further bound as
\begin{equation}
\label{eq33}
\begin{split}
&\sum_{k\neq k' \atop k,k'\in \s}p_{k,\s}p_{k',\s}||\w_{k'}^{(t)}-\w_k^{(t)}||^2\\
\leq& \sum_{k\neq k' \atop k,k'\in \s}p_{k,\s}p_{k',\s} ||\sum_{i=t_0}^{t_0+\tau-1} \eta_i(g_{k'}(\w_{k'}^{(i)},B_{k'}^{(i)})-g_k^{(i)}(\w_k^{(t)},B_k^{(i)}))||^2\\
\leq& \eta_{t_0}^2\tau\sum_{k\neq k' \atop k,k'\in \s}p_{k,\s}p_{k',\s}\sum_{i=t_0}^{t_0+\tau-1}||g_{k'}(\w_{k'}^{(i)},B_{k'}^{(i)})-g_k^{(i)}(\w_k^{(t)},B_k^{(i)})||^2\\
\leq& \eta_{t_0}^2\tau\sum_{k\neq k' \atop k,k'\in \s}p_{k,\s}p_{k',\s}\sum_{i=t_0}^{t_0+\tau-1}[2||g_{k'}(\w_{k'}^{i},B_{k'}^{(i)})||^2 + 2||g_k(\w_k^{(i)},B_k^{(i)})||^2].
\end{split}
\end{equation}According to the Assumption 4, the expectation over Eq.\ref{eq33} can be written as
\begin{equation}
\begin{split}
& \mathbb{E}[\sum_{k\neq k' \atop k,k'\in \s}p_{k,\s}p_{k',\s}||\w_{k'}^{(t)}-\w_k^{(t)}||^2] \\
=& 2\eta_0^2\tau \mathbb{E}[\sum_{k\neq k' \atop k,k'\in \s}p_{k,\s}p_{k',\s}\sum_{i=t_0}^{t_0+\tau-1}(||g_{k'}(\w_{k'}^{i},B_{k'}^{(i)})||^2 + ||g_k(\w_k^{(i)},B_k^{(i)})||^2)]\\
\leq&  2\eta_0^2\tau \mathbb{E}_{\s}[\sum_{k\neq k' \atop k,k'\in \s}p_{k,\s}p_{k',\s}\sum_{i=t_0}^{t_0+\tau-1}2G^2]\\
=& 2\eta_0^2\tau \mathbb{E}_{\s}[\sum_{k\neq k' \atop k,k'\in \s}2p_{k,\s}p_{k',\s}\tau G^2]\\
\leq& 2\eta_0^2\tau \mathbb{E}_{\s}[\sum_{k\neq k' \atop k,k'\in \s}2\tau G^2].
\end{split}
\end{equation}Since there are at most $m(m-1)$ pairs such that $k \neq k'$ in $\s$, we have
\begin{equation}
\begin{split}
& \mathbb{E}[\sum_{k\neq k' \atop k,k'\in \s}p_{k,\s}p_{k',\s}||\w_{k'}^{(t)}-\w_k^{(t)}||^2] \\
\leq& \frac{16\eta_t^2(m-1)\tau^2G^2}{m}\\
\leq& 16\eta_t^2\tau^2G^2.
\end{split}
\end{equation}
\end{proof}

\subsection{Proof of Theorem 1}
\begin{proof}
According to Eq. \eqref{globalUpdate2} we define $\mathcal{H}(\w,t+1)$ as
\begin{equation}
\begin{split}
&\mathcal{H}(\w,t+1) = ||\overline{\w}^{(t+1)}-\w^*||^2 = ||\overline{\w}^{(t)} - \eta_t\overline{g}^{(t)} - \w^*||^2.
\end{split}
\end{equation}
The $\mathcal{H}(\w,t+1)$ can be written as:
\begin{equation}
\begin{split}
&\mathcal{H}(\w,t+1) \\
=& ||\overline{\w}^{(t)} - \w^* - \eta_t\sum_{k \in \s}p_{k,\s}\nabla F_k(\w_k^{(t)}) + \eta_t\sum_{k \in \s}p_{k,\s}\nabla F_k(\w_k^{(t)}) -\eta_t \overline{g}^{(t)}||^2.
\end{split}
\end{equation}
We expand the above formula as follows:
\begin{equation}
\begin{split}
&\mathcal{H}(\w,t+1) \\
=& ||\overline{\w}^{(t)} - \w^* -\eta_t\sum_{k \in \s}p_{k,\s}\nabla F_k(\w_k^{(t)})||^2+ ||\eta_t\sum_{k \in \s}p_{k,\s}\nabla F_k(\w_k^{(t)}) - \eta_t\overline{g}^{(t)}||^2 \\
&+ \underbrace{2\eta_t\langle\overline{\w}^{(t)} - \w^* -\eta_t\sum_{k \in \s}p_{k,\s}\nabla F_k(\w_k^{(t)}),\sum_{k \in \s}p_{k,\s}\nabla F_k(\w_k^{(t)}) - \overline{g}^{(t)}\rangle}_{A_1}.
\end{split}
\end{equation}
Due to the Assumption 3, we have $\mathbb{E}[A_1]=0$ and expand the rest of the formula further:
\begin{equation}
\begin{split}
&\mathcal{H}(\w,t+1) \\
=& ||\overline{\w}^{(t)} - \w^* -\eta_t\sum_{k \in \s}p_{k,\s}\nabla F_k(\w_k^{(t)})||^2+ ||\eta_t\sum_{k \in \s}p_{k,\s}\nabla F_k(\w_k^{(t)}) - \eta_t\overline{g}^{(t)}||^2 +A_1\\
=& ||\overline{\w}^{(t)} - \w^*||^2 + ||\eta_t\sum_{k \in \s}p_{k,\s}\nabla F_k(\w_k^{(t)})||^2 - 2\eta_t\langle\overline{\w}^{(t)}-\w^*,\sum_{k \in S_a^{(t)}}p_{k,\s}\nabla F_k(\w_k^{(t)})\rangle \\
&+ ||\eta_t\sum_{k \in \s}p_{k,\s}\nabla F_k(\w_k^{(t)}) - \eta_t\overline{g}^{(t)}||^2 + A_1.
\end{split}
\end{equation}
Due to the Lemma \ref{l1}, we have that:
\begin{equation}
\begin{split}
&\mathcal{H}(\w,t+1) \\
\leq& ||\overline{\w}^{(t)} - \w^*||^2 + ||\eta_t\sum_{k \in \s}p_{k,\s}\nabla F_k(\w_k^{(t)})||^2 + ||\eta_t\sum_{k \in \s}p_{k,\s}\nabla F_k(\w_k^{(t)}) - \eta_t\overline{g}^{(t)}||^2 \\
&+ \sum_{k \in \s}p_{k,\s}||\overline{\w}^{(t)} - \w_k^{(t)}||^2 + \eta_t^2p_{k,\s}||\nabla F_k(\w_k^{(t)})||^2 - 2\eta_t\sum_{k \in \s}p_{k,\s}(F_k(\w_k^{(t)})-F_k(\w^*)) \\
&- \eta_t\mu\sum_{k \in \s}p_{k,\s}||\w_k^{(t)}-\w^*||^2 +A_1\\
=& ||\overline{\w}^{(t)} - \w^*||^2 + 2\eta_t^2\sum_{k \in \s}p_{k,\s}||\nabla F_k(\w_k^{(t)})||^2 + \sum_{k \in \s}p_{k,\s}||\overline{\w}^{(t)} - \w_k^{(t)}||^2 - 2\eta_t\sum_{k \in \s}p_{k,\s}(F_k(\w_k^{(t)})-F_k(\w^*))\\
&-\eta_t\mu\sum_{k \in \s}p_{k,\s}||\w_k^{(t)}-\w^*||^2 + ||\eta_t\sum_{k \in \s}p_{k,\s}\nabla F_k(\w_k^{(t)}) - \eta_t\overline{g}^{(t)}||^2+A_1\\
\leq& ||\overline{\w}^{(t)}-\w^*||^2 + \underbrace{4L\eta_t^2\sum_{k \in \s}p_{k,\s}(F_k(\w_k^{(t)})-F_k^*)}_{A_2} - \underbrace{2\eta_t\sum_{k \in \s}p_{k,\s}(F_k(\w_k^{(t)})-F_k(\w^*))}_{A_3} \\
&+ \sum_{k \in \s}p_{k,\s}||\overline{\w}^{(t)}- \w_k^{(t)}||^2 -\eta_t\mu\sum_{k \in \s}p_{k,\s}||\w_k^{(t)}-\w^*||^2 + ||\eta_t\sum_{k \in \s}p_{k,\s}\nabla F_k(\w_k^{(t)}) - \eta_t\overline{g}^{(t)}||^2 +A_1. 
\end{split} 
\end{equation}
The difference of $A_2$ and $A_3$ can be written as
\begin{equation}
\begin{split}
&A_2 - A_3 \\
=& (4L\eta_t^2 - 2\eta_t)\sum_{k \in \s}p_{k,\s}F_k(\w_k^{(t)}) - 4L\eta_t^2\sum_{k \in \s}p_{k,\s}F_k^* + 2\eta_t\sum_{k \in \s}p_{k,\s}F_k(\w^*)\\
=& (4L\eta_t^2 - 2\eta_t)\sum_{k \in \s}p_{k,\s}F_k(\w_k^{(t)}) - (4L\eta_t^2-2\eta_t)\sum_{k \in \s}p_{k,\s}F_k^* + 2\eta_t \sum_{k \in \s}p_{k,\s}F_k(\w^*) - 2\eta_t  \sum_{k \in \s}p_{k,\s}F_k^* \\
=& (4L\eta_t^2-2\eta_t)\sum_{k \in \s}p_{k,\s}(F_k(\w_k^{(t)}) - F_k^*) + 2\eta_t\sum_{k \in \s}p_{k,\s}(F_k(\w^*) - F_k^*).
\end{split} 
\end{equation}
Let $v_t = 2\eta_t-4L\eta_t^2 = 2\eta_t(1-2L\eta_t)$, we have that:
\begin{equation}
\begin{split}
&\mathcal{H}(\w,t+1) \\ 
\leq& ||\overline{\w}^{(t)}-\w^*||^2 - v_t\sum_{k \in \s}p_{k,\s}(F_k(\w_k^{(t)}) - F_k^*) + 2\eta_t\sum_{k \in \s}p_{k,\s}(F_k(\w^*) - F_k^*) + \sum_{k \in \s}p_{k,\s}||\overline{\w}^{(t)} - \w_k^{(t)}||^2 \\
&- \eta_t\mu\sum_{k \in \s}p_{k,\s}||\w_k^{(t)}-\w^*||^2 + ||\eta_t\sum_{k \in \s}p_{k,\s}\nabla F_k(\w_k^{(t)}) - \eta_t\overline{g}^{(t)}||^2 + A_1.
\end{split} 
\end{equation}
Due to the Lemma \ref{l2}, we have that:
\begin{equation}
\begin{split}
&\mathcal{H}(\w,t+1) \\
 \leq& ||\overline{\w}^{(t)} - \w^*||^2 +2\eta_t\sum_{k \in \s}p_{k,\s}(F_k(\w^*)-F_k^*) + 2\sum_{k \in \s}p_{k,\s}||\overline{\w}^{(t)} - \w_k^{(t)}||^2 \\
&- v_t(1-\eta_tL)\sum_{k \in \s}p_{k,\s}(F_k(\overline{\w}^{(t)}) - F_k^*) - \eta_t\mu\sum_{k \in \s}p_{k,\s}||\w_k^{(t)}-\w^*||^2 \\
&+ ||\eta_t\sum_{k \in \s}p_{k,\s}\nabla F_k(\w_k^{(t)}) - \eta_t\overline{g}^{(t)}||^2 + A_1.
\end{split} 
\end{equation}
We next solve the expectation over $\mathcal{H}(\w,t+1)$:
\begin{equation}
\begin{split}
&\mathbb{E}[\mathcal{H}(\w,t+1)]=\mathbb{E}[||\overline{\w}^{(t+1)} - \w^*||^2] \\
\leq& \mathbb{E}[||\overline{\w}^{(t)} - \w^*||^2] +\mathbb{E}[2\eta_t\sum_{k \in \s}p_{k,\s}(F_k(\w^*)-F_k^*)] + \mathbb{E}[2\sum_{k \in \s}p_{k,\s}||\overline{\w}^{(t)} - \w_k^{(t)}||^2] \\
&- \mathbb{E}[v_t(1-\eta_tL)\sum_{k \in \s}p_{k,\s}(F_k(\overline{\w}^{(t)}) - F_k^*)] - \mathbb{E}[\eta_t\mu\sum_{k \in \s}p_{k,\s}||\w_k^{(t)}-\w^*||^2] \\
&+ \mathbb{E}[||\eta_t\sum_{k \in \s}p_{k,\s}\nabla F_k(\w_k^{(t)}) - \eta_t\overline{g}^{(t)}||^2] + \mathbb{E}[A_1].\\ 
\end{split} 
\end{equation}
Due to Assumption 3 and $\mathbb{E}[A_1] = 0$, we have
\begin{equation}
\begin{split}
&\mathbb{E}[||\overline{\w}^{(t+1)} - \w^*||^2] \\
\leq& \mathbb{E}[||\overline{\w}^{(t)}-\w^*||^2] + 2\eta_t\mathbb{E}[\sum_{k \in \s}p_{k,\s}(F_k(\w^*)-F_k^*)] + 2\mathbb{E}[\sum_{k \in \s}p_{k,\s}||\overline{\w}^{(t)} - \w_k^{(t)}||^2] \\
&- v_t(1-\eta_tL)\mathbb{E}[\sum_{k \in \s}p_{k,\s}(F_k(\overline{\w}^{(t)})-F_k^*)] - \eta_t\mu\mathbb{E}[\sum_{k \in \s}p_{k,\s}||\w_k^{(t)}- \w^*||^2] + \eta_t^2\sum_{k=1}^Kp_k^2\sigma_k^2.  \\
=& (1-\eta_t\mu)\mathbb{E}[||\overline{\w}^{(t)}-\w^*||^2] + 2\eta_t\mathbb{E}[\sum_{k \in \s}p_{k,\s}(F_k(\w^*)-F_k^*)] + 2\mathbb{E}[\sum_{k \in \s}p_{k,\s}||\overline{\w}^{(t)} - \w_k^{(t)}||^2]
\\
&- v_t(1-\eta_tL)\mathbb{E}[\sum_{k \in \s}p_{k,\s}(F_k(\overline{\w}^{(t)})-F_k^*)] + \eta_t^2\sum_{k=1}^Kp_k^2\sigma_k^2 \\
=& (1-\eta_t\mu)\mathbb{E}[||\overline{\w}^{(t)}-\w^*||^2] + \mathbb{E}[\mathcal{Q}(\w,k,t)] + \eta_t^2\sum_{k=1}^Kp_k^2\sigma_k^2,
\end{split} 
\end{equation}
where $\mathcal{Q}(\w,k,t)$ are defined as follows:
\begin{equation}
\begin{split}
&\mathcal{Q}(\w,k,t) \\
=& 2\eta_t\sum_{k \in \s}p_{k,\s}(F_k(\w^*)-F_k^*)+ 2\sum_{k \in \s}p_{k,\s}||\overline{\w}^{(t)} - \w_k^{(t)}||^2-v_t(1-\eta_tL)\sum_{k \in \s}p_{k,\s}(F_k(\overline{\w}^{(t)})-F_k^*).
\end{split} 
\end{equation}
Note that $S_c^* = \arg\min_{S_c}\sum_{k\in S_c}p_{k,S_c}F_k^*$. Due to the Lemma \ref{l4}, the expectation of the $\mathcal{Q}(\w,k,t)$ can be written as:
\begin{equation}
\begin{split}
&\mathbb{E}[\mathcal{Q}(\w,k,t)] \\
=& -v_t(1-\eta_tL)\mathbb{E}[\sum_{k \in \s}p_{k,\s}(F_k(\overline{\w}^{(t)})-F_k^*)] + 2\eta_t\mathbb{E}[\sum_{k \in \s}p_{k,\s}(F_k(\w^*)-F_k^*]+ 2\mathbb{E}[\sum_{k \in \s}p_{k,\s}||\overline{\w}^{(t)} - \w_k^{(t)}||^2]\\
=& -v_t(1-\eta_tL)\mathbb{E}[\sum_{k \in \s}p_{k,\s}(F_k(\overline{\w}^{(t)})-F_k^*)] + 2\eta_t\mathbb{E}[\sum_{k \in \s}p_{k,\s}(F_k(\w^*)-F_k^*)]+ 32\eta_t^2\tau^2G^2\\
=&-v_t(1-\eta_tL)\mathbb{E}[\sum_{k \in \s}p_{k,\s}(F_k(\overline{\w}^{(t)})-\sum_{k' \in S_c^*}p_{k',S_c^*}F_{k'}^* + \sum_{k' \in S_c^*}p_{k',S_c^*}F_{k'}^* - \sum_{k \in \s}p_{k,\s}F_k^*]\\
&+ 2\eta_t\mathbb{E}[\sum_{k \in \s}p_{k,\s}F_k(\w^*)- \sum_{k' \in S_c^*}p_{k',S_c^*}F_{k'}^* + \sum_{k' \in S_c^*}p_{k',S_c^*}F_{k'}^* -\sum_{k \in \s}p_{k,\s}F_k^*]+ 32\eta_t^2\tau^2G^2\\
=& -v_t(1-\eta_tL)(\mathbb{E}[\sum_{k \in \s} p_{k,\s}F_k(\overline{\w}^{(t)}) - \sum_{k' \in S_c^*}p_{k',S_c^*}F_{k'}^*] + \mathbb{E}[\sum_{k' \in S_c^*}p_{k',S_c^*}F_{k'}^* - \sum_{k \in \s}p_{k,\s}F_k^*]) \\
&+ 2\eta_t(\mathbb{E}[\sum_{k \in \s}p_{k,\s}F_k(\w^*)-\sum_{k' \in S_c^*}p_{k',S_c^*}F_{k'}^*]+\mathbb{E}[\sum_{k' \in S_c^*}p_{k',S_c^*}F_{k'}^*- \sum_{k\in \s}p_{k,\s}F_k^*]) + 32\eta_t^2\tau^2G^2\\
=& -v_t(1-\eta_tL)\mathbb{E}[\sum_{k \in \s}p_{k,\s}F_k(\overline{\w}^{(t)})-\sum_{k' \in S_c^*}p_{k',S_c^*}F_{k'}^*] + 2\eta_t(\mathbb{E}[\sum_{k \in \s}p_{k,\s}F_k(\w^*)-\sum_{k' \in S_c^*}p_{k',S_c^*}F_{k'}^*] \\
&- (2\eta_t-v_t(1-\eta_tL))\mathbb{E}[\sum_{k \in \s}p_{k,\s}F_k^*-\sum_{k' \in S_c^*}p_{k',S_c^*}F_{k'}^*] +32\eta_t^2\tau^2G^2.\\
\end{split} 
\end{equation}
According to the Assumption 5 and Definition 1 and 2, we have
\begin{equation}
\begin{split}
&\mathbb{E}[\mathcal{Q}(\w,k,t)] \\
\leq& -v_t(1-\eta_tL)\mathbb{E}[\varphi(\s,\overline{\w})(F(\overline{\w})-\sum_{k=1}^Kp_{k}F_k^*)] + 2\eta_t\mathbb{E}[\varphi(\s,\w^*)(F*-\sum_{k=1}^K p_kF_k^*)] \\
&+ (2\eta_t-v_t(1-\eta_tL))||\mathbb{E}[\sum_{k \in \s}p_{k,\s}F_k^*-\sum_{k' \in S_c^*}p_{k',S_c^*}F_{k'}^*]||+ 32\eta_t^2\tau^2G^2\\
\leq& -v_t(1-\eta_tL)\mathbb{E}[\varphi(\s,\overline{\w})(F(\overline{\w})-\sum_{k=1}^Kp_kF_k^*)] + 2\eta_t\mathbb{E}[\varphi(\s,\w^*)(F^*-\sum_{k=1}^K p_kF_k^*)] \\
&+ (2\eta_t-v_t(1-\eta_tL))\kappa^2 + 32\eta_t^2\tau^2G^2\\
\leq& -v_t(1-\eta_tL)\varphi_{min}\mathbb{E}[(F(\overline{\w})-\sum_{k=1}^Kp_kF_k^*)] + 2\eta_t\varphi_{max}\mathbb{E}[(F(\w^*)-\sum_{k=1}^Kp_kF_k^*)] + (2\eta_t-v_t(1-\eta_tL))\kappa^2 \\
&+ 32\eta_t^2\tau^2G^2\\
\leq& \underbrace{-v_t(1-\eta_tL)\varphi_{min}\mathbb{E}[(F(\overline{\w})-\sum_{k=1}^Kp_kF_k^*)]}_{A_4} + 2\eta_t\varphi_{max}\Gamma + 6L\eta_t^2\kappa^2+ 32\eta_t^2\tau^2G^2.
\end{split} 
\end{equation}
We can expand the $A_4$ as 
\begin{equation}
\begin{split}
&A_4 \\
=& -v_t(1-\eta_tL)\varphi_{min}\mathbb{E}[(F(\overline{\w})-\sum_{k=1}^Kp_kF_k^*)] \\
=&-v_t(1-\eta_tL)\varphi_{min}\sum_{k=1}^Kp_k(\mathbb{E}[F(\overline{\w})]-F^* + F^*-F_k^*) \\
=& -v_t(1-\eta_tL)\varphi_{min}\sum_{k=1}^Kp_k(\mathbb{E}[F_k(\overline{\w}^{(t)})]-F^*) -v_t(1-\eta_tL)\varphi_{min}\sum_{k=1}^Kp_k(F^*-F_k^*)\\
=& -v_t(1-\eta_tL)\varphi_{min}(\mathbb{E}[F(\overline{\w}^{(t)})]-F^*) - v_t(1-\eta_tL)\varphi_{min}\Gamma\\
\end{split} 
\end{equation}
\begin{equation}
\begin{split}
\leq& -\frac{v_t(1-\eta_tL)\mu \varphi_{min}}{2}\mathbb{E}[||\overline{\w}^{(t)}-\w^*||^2]- v_t(1-\eta_tL)\varphi_{min}\Gamma\\
\leq& -\frac{3\eta_t\mu\varphi_{min}}{8}\mathbb{E}[||\overline{\w}^{(t)}-\w^*||^2] - 2\eta_t(1-2L\eta_t)(1-\eta_tL)\varphi_{min}\Gamma\\
\leq& -\frac{3\eta_t\mu\varphi_{min}}{8}\mathbb{E}[||\overline{\w}^{(t)}-\w^*||^2] - 2\eta_t\varphi_{min}\Gamma + 6\eta_t^2\varphi_{min}L\Gamma.
\end{split} 
\end{equation}
So we have
\begin{equation}
\begin{split}
&\mathbb{E}[\mathcal{Q}(\w,k,t)] \\
=& -\frac{3\eta_t\mu\varphi_{min}}{8}\mathbb{E}[||\overline{\w}^{(t)}-\w^*||^2] + 2\eta_t\Gamma(\varphi_{max}-\varphi_{min}) + \eta_t^2(6\varphi_{min}L\Gamma + 32\tau^2G^2 + 6L\kappa^2).
\end{split} 
\end{equation}
As a result, we have
\begin{equation}
\begin{split}
&\mathbb{E}[||\overline{\w}^{(t+1)}-\w^*||] \\
\leq& [1-\eta_t\mu(1+\frac{3\varphi_{min}}{8})]\mathbb{E}[||\overline{\w}^{(t)}-\w^*||^2] + 2\eta_t\Gamma(\varphi_{max}-\varphi_{min}) + \eta_t^2(6\varphi_{min}L\Gamma + 32\tau^2G^2 + 6L\kappa^2 + \sum_{k=1}^Kp_k\sigma_k^2).
\end{split} 
\end{equation}
By defining $\Delta_{t+1} = \mathbb{E}[||\overline{\w}^{(t+1)}-\w^*||]$, $B = 1+\frac{3\varphi_{min}}{8}$, $C = 6\varphi_{min}L\Gamma + 32\tau^2G^2 + 6L\kappa^2 + \sum_{k=1}^Kp_k\sigma_k^2$, $D = \Gamma(\varphi_{max}-\varphi_{min})$, we have
\begin{equation}
\begin{split}
&\Delta_{t+1} \leq (1-\eta_t\mu B)\Delta_t + \eta_t^2C + \eta_tD.
\end{split} 
\end{equation}
If we set $\Delta_t \leq \frac{\psi}{t+\gamma}$, $\eta_t = \frac{\beta}{t + \gamma}$ and $\beta > \frac{1}{\mu B}$, $\gamma > 0$ by induction, we have
\begin{equation}
\begin{split}
&\psi = \max{\left\{ \gamma||\overline{\w}^1-\w^*||^2,\frac{1}{\beta \mu B - 1}(\beta^2C + D\beta(t+\gamma))\right\}}.
\end{split} 
\end{equation}
Then by the L-smoothness of $F(\cdot)$, 
\begin{equation}
\begin{split}
&\mathbb{E}[F(\overline{\w}^{(t)})] - F^* \leq \frac{L}{2}\Delta_t \leq \frac{L}{2}\frac{\psi}{\gamma + t}.
\end{split} 
\end{equation}
Finally, we complete the proof of Theorem 1:
\begin{equation}
\begin{split}
&\mathbb{E}[F(\overline{\w}^{T})] - F^*\\
\leq& \frac{1}{T+\gamma}\left[ \frac{4L(32\tau^2G^2 + \sum_{k=1}^Kp_k\sigma_k^2) + 24L^2\kappa^2}{3\mu^2 \varphi_{min}}  + \frac{8L^2\Gamma}{\mu^2}+\frac{L\gamma||\overline{\w}^{1} - \w^*||^2}{2}\right] + \frac{8L\Gamma}{3\mu}\left(\frac{\varphi_{max}}{\varphi_{min}}-1\right),
\end{split} 
\end{equation}
where the $T$ means the maximal communication rounds, which satisfies $T = i\tau$ for $i = 1,2,...$ in realistic scenarios.
\end{proof}

\end{document}